\pdfoutput=1

\documentclass[11pt]{article}

\usepackage[final]{acl}

\usepackage{times}
\usepackage{latexsym}

\usepackage[T1]{fontenc}

\usepackage[utf8]{inputenc}

\usepackage{microtype}

\usepackage{inconsolata}

\usepackage{graphicx}
\usepackage{subcaption}
\usepackage{booktabs}
\usepackage{adjustbox}
\usepackage{makecell}
\usepackage{pifont}
\usepackage{multirow}
\usepackage{lipsum}
\usepackage{mwe}
\usepackage{url}
\usepackage{tabularx}
\usepackage{xspace}
\usepackage{booktabs}
\usepackage{amsmath, amssymb}
\usepackage[ruled,vlined,linesnumbered]{algorithm2e}
\usepackage{tcolorbox}
\tcbuselibrary{breakable}
\usepackage{fvextra}
\DefineVerbatimEnvironment{verbatim}{Verbatim}{breaklines=true}

\newcommand{\benchmarkName}{FlashAdventure\xspace}
\newcommand{\methodName}{COAST\xspace}
\newcommand{\numGames}{34\xspace}
\newcommand{\cmark}{\ding{51}}%
\newcommand{\xmark}{\ding{55}}%

\def \ie{\textit{i.e.}, }
\def \eg{\textit{e.g.}, }

\title{\benchmarkName: A Benchmark for GUI Agents Solving Full Story Arcs in Diverse Adventure Games}

\author{
    \quad Jaewoo Ahn$^{*, \dagger, 1,2}$
    \quad Junseo Kim$^{*,1}$
    \quad \textbf{Heeseung Yun}$^{1}$
    \quad \textbf{Jaehyeon Son}$^{\dagger, 2, 3}$ \\ 
    \quad \textbf{Dongmin Park}$^{2}$
    \quad \textbf{Jaewoong Cho}$^{2}$
    \quad \textbf{Gunhee Kim}$^{1}$\\
    $^1$Seoul National University\quad $^2$KRAFTON
    \quad $^3$Georgia Institute of Technology \\
    \texttt{\small \{jaewoo.ahn, junseo.kim, heeseung.yun\}@vision.snu.ac.kr} \\
    \texttt{\small jaehyeon.son@gatech.edu,} \texttt{\small \{dongmin.park, jwcho\}@krafton.com,} \texttt{\small gunhee@snu.ac.kr}\\
    \small{\url{https://ahnjaewoo.github.io/flashadventure}} \\
}

\newcommand{\correspondingfootnote}{
    \let\oldthefootnote=\thefootnote
    \renewcommand{\thefootnote}{}
    \footnotemark
    \footnotetext{$^{*}$Equal contribution.}
    \footnotetext{$\dagger$Work done during an internship at KRAFTON.}
    \let\thefootnote=\oldthefootnote
}

\begin{document}
\maketitle
\begin{abstract}
GUI agents powered by LLMs show promise in interacting with diverse digital environments. Among these, \textit{video games} offer a valuable testbed due to their varied interfaces, with \textit{adventure games} posing additional challenges through complex, narrative-driven interactions. Existing game benchmarks, however, lack diversity and rarely evaluate agents on completing entire storylines. To address this, we introduce \benchmarkName, a benchmark of \numGames Flash-based adventure games designed to test full story arc completion and tackle the \textit{observation-behavior gap}: the challenge of remembering and acting on earlier gameplay information. We also propose CUA-as-a-Judge, an automated gameplay evaluator, and \methodName, an agentic framework leveraging \textit{long-term clue memory} to better plan and solve sequential tasks. Experiments show current GUI agents struggle with full story arcs, while \methodName improves milestone completion by bridging the observation-behavior gap. Nonetheless, a marked discrepancy between humans and best-performing agents warrants continued research efforts to narrow this divide.
\end{abstract}

\correspondingfootnote

\section{Introduction}

Recent advances in Large Language Models (LLMs) and multimodal LLMs have enabled the emergence of graphical user interface (GUI) agents, or Computer-Using Agents (CUA) - autonomous systems that perform diverse digital tasks by controlling mouse and keyboard inputs to interact with visual elements on platforms such as web, mobile, and OS~\citep{nguyen2024guiagentsurvey,tang2025guiagentsurvey}.
This approach holds great promise, as GUIs are ubiquitous in all computing devices humans use in work and daily life.

\begin{figure}[h!]
    \centering
    \includegraphics[width=\columnwidth]{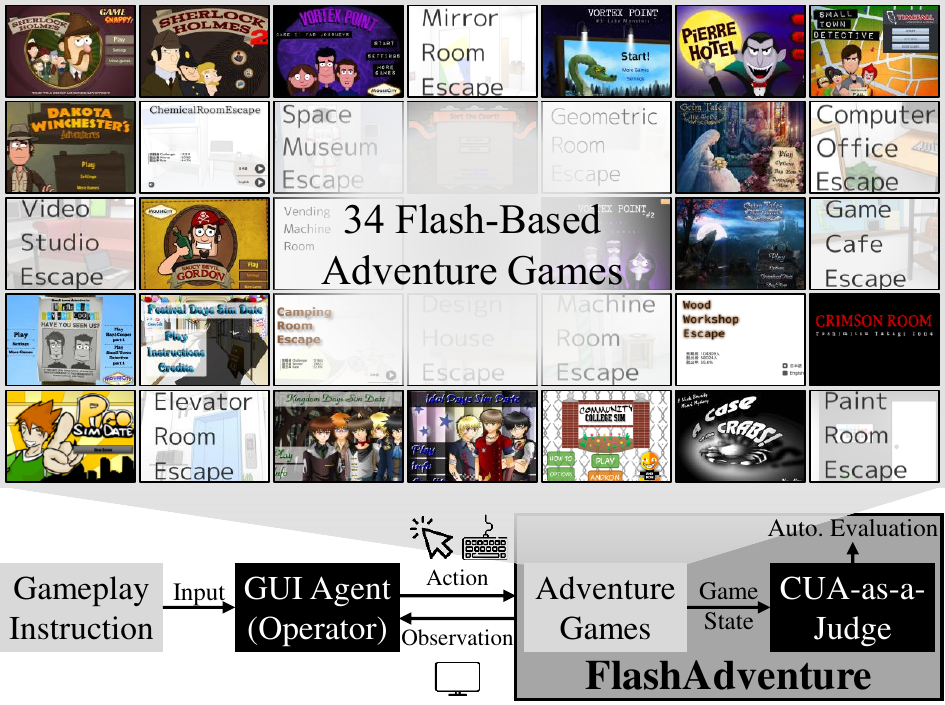}
    \caption{\benchmarkName consists of \numGames Flash-based classic adventure games and supports automatic evaluation of the GUI agent using CUA-as-a-Judge.}
    \label{fig:flashadventure}
\end{figure}

Among various applications of GUI agents, \textit{video gaming} agents play a crucial role.
Video games offer a balanced environment between device user interfaces and real-world perception.
Unlike standard GUIs such as web pages or widgets, game environments feature non-standard, less obvious layouts and complex interaction modes, making them an excellent testbed for evaluating the generalizability of GUI agents~\citep{hu2025surveygameagents}.
In particular, classic \textit{adventure games}\footnote{\url{https://en.wikipedia.org/wiki/Adventure_game}.}, characterized by narrative-driven exploration, introduce additional complexity through diverse visual layouts and interactions, 
such as managing inventories, engaging in multi-turn dialogues with branching outcomes, and performing context-sensitive actions.
Moreover, understanding the story arc requires conceptual reasoning and lateral thinking~\citep{de1970lateral}, as players should interweave collected items and narrated problems with creativity.

Recent studies of GUI agents in video games have explored several adventure game characteristics, such as performing low-level tasks (\eg renaming a card deck)~\citep{hu2024ootb}, executing sequences of actions (\eg combatting a boss monster)~\citep{chen2024varp}, or completing specific missions (\eg searching barn)~\citep{tan2024cradle}.
While such capabilities allow for meaningful gameplay, a critical gap remains in both (1) the diversity of tasks and (2) the completion of full story arc.
Although these agents excel in specific game tasks, it is unclear whether they can generalize to diverse scenarios.
Moreover, due to the extremely long story arcs in AAA games or the lack of clear narrative conclusions often found in MMORPG-ish games, the ability of GUI agents to complete entire story arcs has yet to be verified.

\begin{figure}[t!]
    \centering
    \includegraphics[width=\columnwidth]{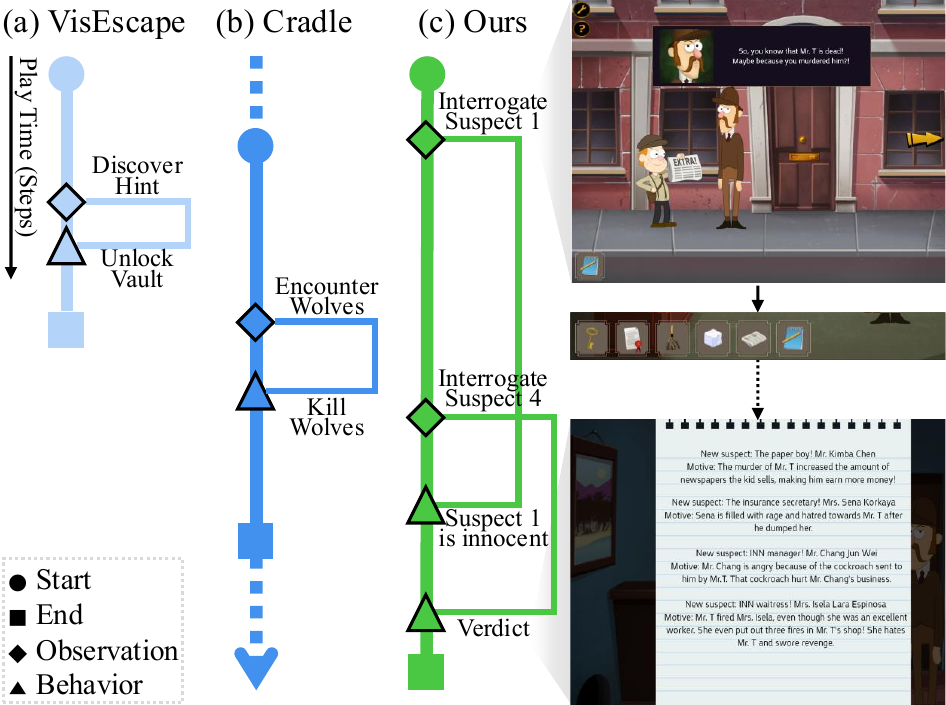}
    \caption{Comparison of gameplay progression across (a) VisEscape~\citep{lim2025visescape}, (b) Cradle~\citep{tan2024cradle}, and (c) \benchmarkName. Prior benchmarks focus on short-term objectives or include short story arcs, limiting their ability to fully evaluate agents’ capacity to manage the long-term observation-behavior gap. In contrast, \benchmarkName emphasizes completion of full story arcs involving long-term objectives, exemplified by suspect interrogations leading to a verdict.}
    \label{fig:motivation}
    \vspace{-1.0em}
\end{figure}

To address these limitations, we introduce \textbf{\benchmarkName}, a new benchmark based on \numGames adventure games to evaluate GUI agents solving full story arcs from the start to finish.
To assess agents in games with self-contained, compact story arcs, we use \textit{Flash} games\footnote{Flash games refer to browser-based games developed using the Adobe Flash platform.} (See Figure~\ref{fig:flashadventure}), which offer compact playtime (approximately one hour per game, compared to over 30 hours for AAA games like \textit{Red Dead Redemption 2}), and are free to play.
A key challenge in this context is the ``observation-behavior gap'', which refers to the \textit{time lag} between when an agent \textit{observes information} and when it can \textit{act upon it}.
As shown in Figure~\ref{fig:motivation}, adventure games require agents to manage long-term time lags, such as interrogating a suspect and later discovering their innocence.
These long-term dependencies are crucial when solving full story arcs.
Tolman's theory on latent learning~\citep{tolman1930introduction, tolman1948cognitive} suggests that humans can retrieve and apply clues after a long delay, which can also be explored in agents to assess whether similar emergent behaviors occur.

Based on our benchmark, we propose two technical components: \textbf{CUA-as-a-Judge} and \textbf{\methodName}.
While prior studies of GUI agents in video games relied on manual assessment, we address this limitation by introducing CUA-as-a-Judge, a new \textit{judge agent}.
This agent acts as an oracle with access to predefined success milestones for each game and actively interacts with the game environment to verify whether these milestones have been achieved.
Additionally, to address observation-behavior gap issues in \benchmarkName, we propose a new agentic framework, \methodName (Clue-Oriented Agent for Sequential Tasks), which leverages long-term \textit{clue memory} by proactively maintaining it and applying it during the problem-solving stage. 

Using \benchmarkName, we evaluate multiple GUI agents and empirically show that they greatly struggle to complete full story arcs due to weak planning capabilities, limited visual perception in non-standard layouts, and the lack of lateral thinking.
In contrast, \methodName enhances the planning by effectively managing clue memory, thereby bridging the observation-behavior gap and improving problem-solving capabilities.
However, all GUI agents still show a significant gap compared to human performance, demonstrating the need for further improvement in associating time-dependent observations with the full story arcs.

\begin{table*}[h!]
\centering
\begin{adjustbox}{width=\linewidth}
\begin{tabular}{lcccccl}
\toprule
\makecell[l]{\textbf{Benchmark /}\\\textbf{Framework}} & \textbf{\# Games} & 
\textbf{Environment} & \textbf{Free?} & \makecell{\textbf{Automatic}\\\textbf{Evaluation}} & \makecell{\textbf{Complete}\\\textbf{Story Arc}} & \textbf{Featured Games} \\ 
\midrule
\multicolumn{7}{l}{\textbf{Code/API-based}} \\
PokeLLMon~(\citeyear{hu2024pokellmon}) & 1 & API & \cmark & \cmark & \xmark & Pokémon \\
TextStarCraft II~(\citeyear{ma2024textstarcraft2}) & 1 & API & \cmark & \cmark & \xmark & StarCraft II \\
MineDojo~(\citeyear{fan2022minedojo}) & 1 & API \& Screen & \cmark & \cmark & \xmark & Minecraft \\
BALROG~(\citeyear{paglieri2025balrog}) & 6 & API \& Screen & \cmark & \cmark & \xmark & MiniHack, NLE, Baba Is AI, Crafter, BabyAI, TextWorld \\
LVLM-Playground~(\citeyear{wang2025lvlmplayground}) & 6 & API \& Screen & \cmark & \cmark & \xmark & TicTacToe, Reversi, Sudoku, Minesweeper, Gomoku, Chess \\
V-MAGE~(\citeyear{zheng2025vmage}) & 5 & API \& Screen & \cmark & \cmark & \xmark & FlappyBird, RaceGame, Super Mario, PongGame, Tempest Run \\
VisEscape~(\citeyear{lim2025visescape}) & -$^*$ & API \& Screen & \cmark & \cmark & \cmark & $^*$Room escape game created for research, instead of adapting existing ones \\
Orak~(\citeyear{park2025orak}) & 12 & API \& Screen & \xmark & \cmark & \xmark & 2 Games $\times$ 6 Genres (Action, Adventure, RPG, Simulation, Strategy, Puzzle) \\
\midrule
\multicolumn{7}{l}{\textbf{Pixel/Screenshot-based}} \\
OOTB~(\citeyear{hu2024ootb}) & 2 & Screen Only & \cmark & \xmark & \xmark & Hearthstone, Honkai: Star Rail \\
VARP~(\citeyear{chen2024varp}) & 1 & Screen Only & \xmark & \xmark & \xmark & Black Myth: Wukong (AAA game) \\
Cradle~(\citeyear{tan2024cradle}) & 4 & Screen Only & \xmark & \xmark & \xmark & RDR2 (AAA game), Stardew Valley, Cities: Skylines, Dealer's Life 2 \\
\textbf{\benchmarkName} & \textbf{\numGames} & Screen Only & \cmark & \cmark & \cmark & \makecell[l]{Classic Adventure Games (Mystery/Detective, Hidden Object,\\Room Escape, Visual Novel, Life/Management Simulation)} \\
\bottomrule
\end{tabular}%
\end{adjustbox}
\caption{Overview of video game benchmarks. 
\textit{Complete Story Arc} indicates whether the benchmark evaluates an agent's ability to complete a self-contained story arc from beginning to end.
Our \benchmarkName evaluates agents on completing full story arcs in diverse adventure games.
} 
\vspace{-1.0em}
\label{table:game_benchmarks}
\end{table*}

\section{Related Work}

\textbf{GUI Agents.}
Multimodal LLMs enabled GUI agents to perform tasks on various digital platforms~\citep{nguyen2024guiagentsurvey}.
GUI agents require explicit action planning and execution, necessitating  \textit{GUI grounding} problem~\citep{cheng2024seeclick,wu2025osatlas} to map language instructions to specific GUI elements and executions.
Another key problem is \textit{input modality selection}: whether GUI agent should rely on platform-specific structured data (\eg, DOM trees for web, accessibility trees for OS, view hierarchies for mobile)~\citep{deng2023mindweb,zhou2024webarena,li2020pixelhelp}, vision-only input (screenshots)~\citep{gou2025uground,hong2024cogagent,rawles2023androidinthewild}, or a hybrid combination of both~\citep{he2024webvoyager,koh2024visualwebarena,furuta2024webgum,xie2024osworld}.
Although numerous benchmarks have been proposed to evaluate GUI agents, they focus on standardized templates, restricting the range of interaction behaviors and visual structures that agents encounter~\citep{fan2024screenpr,sun2025guixplore}.

\noindent
\textbf{Video Game Benchmarks.}
Early gameplaying agents relied on reinforcement learning (RL) for solving relatively simple, rule-based environments~\citep{bellemare2013atari,brockman2016openaigym,silver2016alphago}. Recent attention has shifted toward tackling complex video games via LLM-based agents~\citep{wu2024smartplay,raad2024sima,kanervisto2025wham,yuan2025turnaboutllm,zhang2025videogamebench}.
To evaluate game-solving capabilities, several game benchmarks have emerged, broadly categorized into two types.

(1) \textit{Code/API}-based benchmarks provide structured representations of game states or internal reward signals through direct API interactions (\eg, Pokémon~\citep{hu2024pokellmon}, StarCraft II~\citep{ma2024textstarcraft2}). Certain variants augment these interactions with visual inputs, such as screenshots (\eg, Minecraft~\citep{fan2022minedojo}, video game collections~\citep{wang2025lvlmplayground,paglieri2025balrog,hu2025lmgamebench}), yet the fundamental interaction paradigm remains reliant on APIs rather than GUI manipulations (see Table~\ref{table:game_benchmarks}). While such benchmarks guarantee precise and immediate feedback, they are inherently limited to the games that expose accessible APIs, necessitate substantial preliminary setup efforts (\eg modding\footnote{\url{https://en.wikipedia.org/wiki/Modding}} or reverse engineering) to integrate new games.

(2) \textit{Pixel/screenshot}-based benchmarks adopt a more realistic vision-action interface, 
where agents perceive raw visual input (\ie screenshot) and interact via actions such as clicks or keystrokes.
While this paradigm holds promise for general-purpose GUI agent evaluation, existing studies~\citep{chen2024varp, tan2024cradle, chen2025combatvla} often rely on \textit{buy-to-play} AAA games, which limit accessibility and reproducibility, and require manual evaluation, making broader assessments difficult.

Overall, existing game benchmarks lack task and game diversity, featuring at most 12 games (see Table~\ref{table:game_benchmarks}), making it difficult to evaluate agents across a wide range of gaming scenarios beyond specific tasks or a limited set of games.
Furthermore, except for room escape games, none of these benchmarks test agents on complete story arcs, either because AAA games have extensive length~\citep{chen2024varp} or because some games lack clear endings~\citep{hu2024ootb,tan2024cradle}.
Although room escape games evaluate agents on complete story arcs, they fall short in assessing the extensive observation-behavior gap due to their substantially low average human playtimes (\eg 5.7 steps for \citet{wang2025escapecraft}, 52.8 for \citet{lim2025visescape}, and 257.8 for \citet{qian2025escapebench}), as illustrated in Figure~\ref{fig:motivation}.
This limitation stems from the fact that the games were designed specifically for research rather than adapted from existing games for entertainment purposes.

In addition, we further provide a discussion of prior text adventure games and benchmarks in Appendix~\ref{subsec:appendix_related_work}.


\section{The \benchmarkName Benchmark}

We introduce \textbf{\benchmarkName}, a new benchmark comprising \numGames diverse Flash-based classic adventure games that evaluates agents' ability to solve full story arcs.

\subsection{Problem Formulation}
We cast \benchmarkName gameplay as a partially-observable Markov decision process (POMDP) $\langle \mathcal{S},\mathcal{A},\mathcal{O},\Omega,\mathcal{T}, \mathcal{R}\rangle$.
The hidden state space $\mathcal{S}$ denotes the internal configuration of the game runtime (\eg memory, object stacks), while the observation space $\mathcal{O}$ is the RGB frame buffer rendered each step.
A deterministic rendering function $\Omega:\mathcal{S}\rightarrow\mathcal{O}$ returns the visible frame $o_t=\Omega(s_t)$ presented to the agent.
The action set $\mathcal{A}$ comprises low-level GUI inputs (\eg $\texttt{left-click(x,y)}$; see Table~\ref{tab:action_space} in Appendix~\ref{subsec:appendix_problem_formulation} for the complete set) that the agent can issue at every step.
Upon executing $a_t\in\mathcal{A}$, the game engine advances according to the transition kernel $\mathcal{T}:\mathcal{S}\times\mathcal{A}\rightarrow\mathcal{S}$, yielding the next hidden state $s_{t+1}$ and its observation $o_{t+1}$.
A GUI agent policy $\pi_{\theta}$ conditions on a task query $q$ (\eg ``escape the room within 1,000 steps''), and the complete interaction history to output the next action $a_t$, thereby generating the trajectory $\tau=(o_1,a_1,\dots,o_T,a_T)$.
Finally, a reward function $\mathcal{R} : \mathcal{S} \times \mathcal{A} \to \{0, 1\}$ produces a binary success indicator at the end of an episode, which is directly observable to the agent: $\mathcal{R}(s_T, a_T) = 1$ if the agent completes the task successfully, and 0 otherwise.

Separately, within the environment, we define a milestone reward function $\mathcal{R}_{\mathrm{m}} : \mathcal{S} \times \mathcal{A} \to \mathbb{R}$ that assigns intermediate rewards based on progression milestones.
Unlike $\mathcal{R}$, the values of $\mathcal{R}_{\mathrm{m}}$ are not directly observable by the agent.

\subsection{Game Selection}
\label{subsec:game_selection}
We utilize the \textit{FlashPoint Archive}\footnote{\url{https://flashpointarchive.org/}} that enables secure playback of Flash-based browser games.
We specifically focus on \textit{classic adventure games} characterized by narrative-driven exploration.
From the extensive library of Flash-based adventure games, we selected \numGames games based on the following criteria:
\begin{enumerate}
    \item \textit{Free-to-play} games covering diverse subgenres (\eg mystery/detective, hidden object, room escape, visual novel, simulation).
    \item Games that emphasize reasoning over reaction speed by avoiding strict time limits and rapid-action requirements.
    \item Games with validated human walkthroughs and clearly defined progression milestones, excluding fully open-ended narratives. All games are solvable within at most 1-2 hours by human players, ensuring agents can complete the full story arcs within practical evaluation timeframes.
\end{enumerate}

The resulting \numGames game collection makes \benchmarkName the largest among existing video game benchmarks (see Table~\ref{table:game_benchmarks}).
Appendix~\ref{subsec:appendix_game_selection} lists all games and includes a representative human walkthrough for each subgenre. Appendix~\ref{subsec:analysis_game_diversity} further analyzes inter- and intra-subgenre diversity, demonstrating that \benchmarkName provides substantial diversity across games.

\subsection{Human GamePlay}
\label{subsec:human_game_play}
To validate that the selected games are neither too easy nor too difficult to complete full story arcs within feasible timeframes, we collected a set of human gameplay demonstrations.
Specifically, we recruited 13 participants to play through all \numGames games.
Detailed procedures for collecting human gameplay demonstrations are described in Appendix~\ref{subsec:appendix_human_game_play}.

As summarized in Table~\ref{tab:detailed_results_34} (in Appendix~\ref{subsec:appendix_experimental_results}), participants completed full story arcs with an average of 1,142 steps, an average playtime of 26 minutes, and a success rate of 97.1\%, demonstrating that the games can be fully solved within practical time limits.
Compared to prior room escape benchmarks, which require 5.7-257.8 steps to complete the story arcs, our selected games require substantially longer playtime (1,142 steps), reflecting greater complexity.

Moreover, we observed several instances of the long-term observation-behavior gap during gameplay.
As shown in Figure~\ref{fig:obgap_qual} (Appendix~\ref{subsubsec:appendix_obgap_stat}), players exhibit substantial step gaps of 99 and 421 between acquiring items and receiving rewards, demonstrating their ability to manage extended dependencies.
Appendix~\ref{subsubsec:appendix_obgap_stat} reports statistics of the observation-behavior gap, with an average of 251.1 steps, indicating its substantial magnitude.

\subsection{Evaluation}
\label{subsec:evaluation}
\textbf{Defining milestones and scores.}
To systematically evaluate gameplay performance in \benchmarkName, it is essential to clearly define milestones and subtasks reflecting agents' progression towards the narrative end-goal. 
Following prior video game benchmarks using milestones to measure progress~\citep{chen2024varp,tan2024cradle,lim2025visescape}, we define a set of milestones for each game.
To ensure the validity of the defined milestones, three of the authors independently review the full walkthroughs of each game and finalize the milestone list by filtering out those without consensus.
Specifically, for the 27 non-simulation-based  games, we define a \textit{discrete} number of milestones (average: 6.7, maximum: 12, minimum: 4). 
For the 7 simulation-based games, we dispense with discrete milestones and instead use the continuous score shown in the game’s head-up display (HUD).
This score, which can be revealed by clicking the score icon when necessary, is normalized by the maximum attainable value.
We provide the full list of milestones along with a brief introduction to the game in Appendix~\ref{subsec:appendix_evaluation}.

\noindent
\textbf{Evaluation metrics.}
We adopt standard evaluation metrics commonly employed in benchmarks for games and embodied agents~\citep{chen2024varp,tan2024cradle,lim2025visescape}:

\begin{itemize} 
    \item \textbf{Success Rate}: A binary metric $\mathcal{R}(s_T, a_T)$ indicating whether the final narrative goal (\eg mystery solving, room escape) is successfully achieved (1) or not (0).
    \item \textbf{Milestone Completion Rate}: The proportion of milestones completed $\mathcal{R}_{\mathrm{m}}(s_T, a_T)$ at the point when an agent reaches the maximum allowed steps (\eg, 1,000). For simulation-based games, this metric is represented by normalizing the continuous score achieved at the maximum step count against the maximum attainable score.
    \item \textbf{Steps}: The total number of actions taken by the agent ($t$), capped by a predefined maximum number.
\end{itemize}

\subsection{CUA-as-a-Judge}
\label{subsec:cua_as_a_judge}
Existing pixel/screenshot-based video game benchmarks for evaluating progression milestones of GUI game agents lack automatic evaluation methods and have relied on manual assessments by human annotators~\citep{chen2024varp, tan2024cradle}.

To overcome this limitation, we introduce a novel automatic evaluation framework, \textit{CUA-as-a-Judge}, implemented using Claude-3.7-Sonnet computer-use~\citep{anthropic2024claudecomputeruse}.
This \textit{judge} agent functions as an oracle with direct access to success milestones for each game, simulating a human judging process (\eg manually checking puzzle completion or character progress).
After a game agent $\pi_\theta$ finishes gameplay (either by successfully achieving the final goal or reaching the maximum allowed number of steps), the CUA-as-a-Judge resumes from the game's final state $s_T$.
The judge interacts directly with the game environment by executing actions $a_{T+j}$ (\eg $\texttt{mouse-click}(x,y)$, $\texttt{key-type}(\textit{key})$) and observing the resulting observations $o_{T+j}$, thereby verifying the accomplishment of necessary milestones. 
For instance, in Figure~\ref{fig:cua_as_a_judge_example} (right) in Appendix~\ref{subsec:appendix_cua_as_a_judge}, CUA-as-a-Judge can click on the notebook to reveal the number of discovered suspects, directly confirming milestone progression.

Specifically, to determine the degree of completion, the judge sequentially checks whether each of the $N_\text{m}$ predefined milestones has been achieved, starting from the first.
If a milestone is not satisfied, the evaluation halts, and the completion score is computed as $K_{\mathrm{m}}/N_{\mathrm{m}}$, where $K_{\mathrm{m}}$ is the number of milestones successfully verified.
This process continues until failure or full completion of all milestones.
Note that in other cases, milestone verification can instead be performed in a single pass by counting cumulative game states such as collected items or unlocked locations.
Further implementation details on evaluation strategy are provided in Appendix~\ref{subsec:appendix_cua_as_a_judge}.

\noindent
\textbf{Evaluation agreement vs. human judge.}
We evaluate the reliability of CUA-as-a-Judge by comparing its judgments with human judgments across all \numGames games.
For each game, we sampled 8-9 episodes, resulting in a total of 300 samples.
Our comparison shows a high agreement, with an accuracy of 94.00\%, Spearman correlation of 0.9912, and Pearson correlation of 0.9999.
The judge performed particularly well on step-by-step milestone verification, while showing some limitations with counting-based milestones that require assessing multiple items at once.
See Appendix~\ref{subsec:appendix_cua_as_a_judge} for validation details.

\section{Approach}
\label{sec:approach}

\begin{figure*}[h!]
    \centering
    \includegraphics[width=\linewidth]{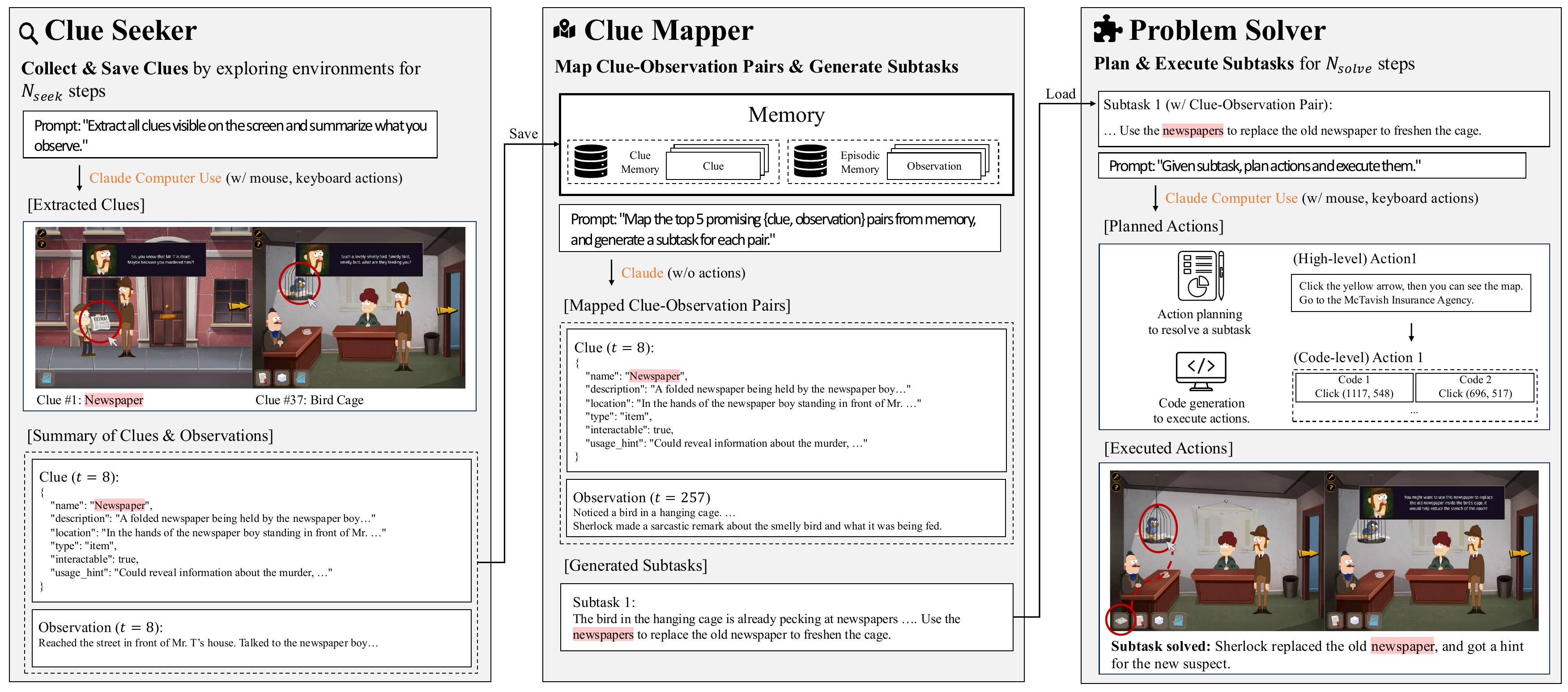}
    \caption{Overview of \methodName Framework with Seek-Map-Solve Cycle.}
    \label{fig:coast}
\end{figure*}

As shown in \S\ref{subsec:human_game_play}, it is essential to address long-term observation-behavior gap in \benchmarkName.
To this end, we propose the \methodName (\textit{Clue-Oriented Agent for Sequential Tasks}) framework, which leverages long-term \textit{clue memory} that stores relevant clues for future use.
Then, \methodName guides agents to actively seek clues and generate diverse subtasks with plausible hypotheses for solving complex tasks by following a \textit{Seek-Map-Solve} cycle (see Figure~\ref{fig:coast} and Algorithm~\ref{alg:method_coast}):

\begin{algorithm}[h!]
\footnotesize
\caption{\textsc{\methodName} Framework with Seek-Map-Solve Cycle.}
\label{alg:method_coast}

\KwIn{task query $q$, maximum step $T$}

\KwData{$\mathcal{M}$ (clue memory), $\tau$ (trajectory), $\mathcal{G}_R$ (resolved-goal set)}

\DontPrintSemicolon
$\mathcal{M} \leftarrow \varnothing$, $\tau \leftarrow \varnothing$, $\mathcal{G}_R \leftarrow \varnothing$, $t \leftarrow 0$\; 

\While{$t < T$}{
  \tcp*[l]{\textbf{1. Clue Seeker}}
  \For{$i = 1$ \KwTo $N_{\text{seek}}$}{
      \If{$t \ge T$}{\textbf{break}}
      Observe frame $o_t$\;
      $a_t\ \leftarrow \pi^{\mathrm{seek}}_\theta(o_t, \mathcal{M}, q)$\;
      $\Delta\mathcal{M} \leftarrow$ Execute action $a_t$\;
      $\mathcal{M} \leftarrow \mathcal{M} \cup \Delta\mathcal{M}$\;
      $\tau \leftarrow \tau \oplus (o_t, a_t)$\tcp*{append $(o_t, a_t)$ to $\tau$}\;
      $t \leftarrow t + 1$\;
  }
  \If{$t \ge T$}{\textbf{break}}
  \tcp*[l]{\textbf{2. Clue Mapper}}
  $\mathcal{G} \leftarrow \pi^{\mathrm{map}}_\phi(\mathcal{M}, \tau, q)$\;
  $\mathcal{G} \leftarrow \mathcal{G} \setminus \mathcal{G}_R$\tcp*{filter out resolved goals}
  \uIf{$\mathcal{G} = \varnothing$}{
      \textbf{continue} \tcp*{restart Seek block}
  }\Else{
      \tcp*[l]{\textbf{3. Problem Solver}}
    \ForEach{$g \in \mathcal{G}$}{
    \For{$j = 1$ \KwTo $N_{\text{solve}}$}{
      \If{$t \ge T$}{\textbf{break}}
      $a_t \leftarrow \pi^{\mathrm{solve}}_\psi(o_t, g, q)$\;
      $\mathtt{success} \leftarrow $ Execute action $a_t$\;
      $\tau \leftarrow \tau \oplus (o_t, a_t)$\;
      \If{$\mathtt{success}$}{
          $\mathcal{G}_R \leftarrow \mathcal{G}_R \cup \{g\}$\;
      }
      $t \leftarrow t + 1$\;
    }
    }
  }
}
\end{algorithm}

\noindent
\textbf{Step 1: Clue Seeking.}
The \textbf{Clue Seeker} module ($\pi^{\mathrm{seek}}_\theta$) explores the environment for $N_\text{seek}$ steps to collect potential clues (\eg ``A folded newspaper held by the boy.''), rather than solving tasks directly. 
All gathered information is stored in a shared clue memory $\mathcal{M}$.

\noindent
\textbf{Step 2: Clue-Observation Mapping.}
The \textbf{Clue Mapper} ($\pi^{\mathrm{map}}_\phi$) analyzes the accumulated memory and trajectory to identify at most $K$ promising pairs of \textit{clues} and corresponding \textit{observations} $(o_t)$ at specific time steps.
Based on these pairs, it generates subtasks (\eg ``Use the newspaper to replace the old newspaper to freshen the bird cage.'') that represent plausible hypotheses, thus forming the goal candidate set $\mathcal{G} = \{g_1, g_2, \ldots, g_K\}$, where each $g_k$ represents a goal defined by ($\mathcal{M}_k, o_{t_k}, \text{subtask}_k$).

\noindent
\textbf{Step 3: Problem Solving.}
The \textbf{Problem Solver} ($\pi^{\mathrm{solve}}_\psi$) executes the proposed subtasks for $N_\text{solve}$.
Completed goals are stored in a resolved set $\mathcal{G}_R$ to avoid duplication.

The agent cycles through clue seeking, subtask generation, and problem solving until the step limit $T$ is reached, continually updating memory and filtering resolved goals.
Note that the clue memory $\mathcal{M}$ has no strict size limit.
It does not cause memory issues storing all clues during gameplay, since each clue is small (\eg a few dozen tokens).
On average, 150 clues $\times$ 85 tokens per clue yields about 12.8K tokens, which is well below the maximum context length of modern LLMs (\eg Claude-3.7-Sonnet supports 200K tokens).


\begin{table*}[t]
\centering
\resizebox{\linewidth}{!}{
\begin{tabular}{lll|ccc}
\toprule
\textbf{Model} & \textbf{GUI Grounding / Action Execution} & \makecell[l]{\textbf{Agentic}\\\textbf{Framework}} & \makecell{\textbf{Success}$\uparrow$\\(\%)}  & \makecell{\textbf{Milestone}$\uparrow$\\(\%)} &\textbf{\# Steps} \\
\midrule
\multirow{1}{*}{GPT-4o} 
 & UGround-V1-7B / \texttt{pyautogui} & Cradle & 0.00 & 4.56 & 1000.0 \\
\midrule
\multirow{3}{*}{\makecell[l]{Claude-3.7-Sonnet}} 
 & UGround-V1-7B / \texttt{pyautogui} & Cradle & 0.00 & 6.59 & 1000.0 \\
\cmidrule{2-6}
 & \multirow{2}{*}{Claude-3.7-Sonnet / \texttt{pyautogui}} & Cradle & 0.00 & 10.60 & 1000.0 \\
 & & Agents S2 & 0.00 & 1.20 & 1000.0 \\
\midrule
\multicolumn{3}{c|}{UI-TARS-1.5-7B} & 0.00 & 6.93 & 1000.0 \\
\multicolumn{3}{c|}{OpenAI CUA} & \textbf{5.88} & 15.39 & 954.1 \\
\multicolumn{3}{c|}{Claude-3.7-Sonnet Computer-Use} & 0.00 & \underline{17.11} & 992.4 \\
\multicolumn{3}{c|}{Claude-3.7-Sonnet Computer-Use + \methodName \textbf{(Ours)}} & \textbf{5.88} & \textbf{19.89} & 966.8 \\
\midrule
\multicolumn{3}{c|}{Human Performance (max 1,000 steps)} & 50.98 & 78.98 & 815.5 \\
\multicolumn{3}{c|}{Human Performance (unlimited)} & 97.06 & 100.00 &  1142.0 \\
\bottomrule
\end{tabular}
}
\caption{Comparison of different GUI agents across all \numGames video games.}
\label{tab:performance_comparison_full}
\end{table*}

\section{Experiments}

\subsection{Baseline Agents}

We evaluate five representative GUI agents, grouped by the degree of modularization between the backbone VLM, the GUI Grounding/action-execution layer, and the agentic framework.

\noindent
\textbf{End-to-end agents.}
(1) Claude-3.7-Sonnet Computer-Use~\citep{anthropic2024claudecomputeruse} and (2) OpenAI CUA~\citep{openai2025operator} are proprietary agents that integrate perception, planning, and action execution within a unified architecture.
(3) UI-TARS-1.5-7B~\citep{qin2025uitars} is an open-source alternative, built upon Qwen-2-VL 7B~\citep{wang2024qwenvl2}.

\noindent
\textbf{Modular agents.}
In contrast, (4) Cradle~\citep{tan2024cradle} and (5) Agent S2~\citep{agashe2025agents2} exemplify modular agents that explicitly separate each module.
We adopt GPT-4o~\citep{openai2024gpt4o} and Claude-3.7-Sonnet~\citep{anthropic2024claude37sonnet} as backbone VLMs.
For GUI Grounding/action-execution, we utilize either UGround-V1-7B~\citep{gou2025uground} or Claude-3.7-Sonnet combined with \texttt{pyautogui} (\ie Python scripts that control the mouse and keyboard.). We exclude GPT-4o as it frequently failed to localize UI elements.

The agentic framework for Cradle comprises six modules: \textit{Information Gathering} process multimodal input, \textit{Self-Reflection} reconsiders past trajectories, \textit{Task Inference} selects the subsequent task, \textit{Skill Curation} generates relevant skills for a given task, \textit{Action Planning} determines executable actions, and \textit{Memory} stores and retrieves past experiences and skills.
Agent S2 extends this modular design by further decoupling grounding and planning, routing each sub-goal to a \textit{Mixture-of-Grounding} ensemble and employing \textit{Proactive Hierarchical Planning} that refines action sketches as new observations arrive.

\subsection{Experimental Settings}
 We set the maximum number of action steps per agent to 1,000.
 For the main experiments, we conducted evaluations on all \numGames games using seven agents.
In addition, we report detailed results for each game in Table~\ref{tab:detailed_results_34} and Table~\ref{tab:detailed_results_13} (Appendix~\ref{subsec:appendix_experimental_results}).
Please refer to Appendix~\ref{subsec:appendix_implementation_details} for further details.

\subsection{Experimental Results}
As shown in Table~\ref{tab:performance_comparison_full} and Table~\ref{tab:detailed_results_34} in Appendix~\ref{subsec:appendix_experimental_results}, all agents exhibit near-zero success rates and significantly lower milestone completion rates compared to humans.
Even the highest-performing agents, measured by success rate, only completed two hidden-object games (\ie the Grim Tales series) and failed to complete any other games.
We identify three common failure patterns that contribute to the agents’ inability to complete full story arcs, as follows:

\noindent
\textbf{Weak planning capability.}
Agents exhibit repetitive behaviors, such as revisiting the same locations or repeating identical actions, that consume excessive turns.
This arises from insufficient planning abilities and limited memory, preventing agents from effectively leveraging past interactions to guide their actions.
As a result, even when agents have relevant clues, they often fail to generate appropriate subtask plans, reflecting a significant gap between their observations and behavior.

\noindent
\textbf{Poor visual perception.}
Agents often fail to correctly interpret non-standard layouts, resulting in inappropriate interactions.
For example, hidden object games (see Figure~\ref{fig:walkthrough_grim_tales} in Appendix~\ref{subsec:appendix_game_selection}), which have simple structures solvable through visual recognition, still pose challenges for most baselines.

\noindent
\textbf{Deficient lateral thinking.}
When subtasks are not explicitly defined, flexible problem-solving in uncertain environments becomes essential; however, agents often exhibit inflexible thinking, which ultimately reveals weaknesses in planning and reasoning.
For instance, in an escape room game where the goal is to find clues to get out, an agent might focus only on the final goal and keep roaming around the door, missing the important clues nearby.

\begin{figure}[t!]
    \centering
    \includegraphics[width=\columnwidth]{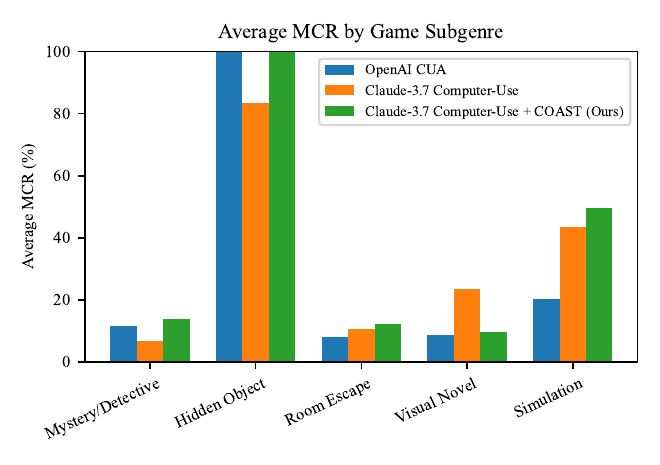}
    \caption{Comparison of average milestone completion rates (MCR) across different game subgenres for three GUI agents.}
    \label{fig:mcr_by_subgenre}
\end{figure}

In contrast, \methodName improves planning by incorporating clue memory: it first seeks to gather as many clues as possible, then explores the available situations and uses this information for subtask planning.
This approach helps to address the observation-behavior gap and experimentally demonstrates some degree of lateral thinking.
As a result, it achieves the highest success and milestone completion rates among the baselines, improving success rate by 5.88 percentage points and milestone completion by up to 2.78 percentage points compared to Claude-3.7-Sonnet Computer-Use.
However, in visual novels, our method does not consistently outperform the baseline Claude-3.7-Sonnet Computer Use, as shown in Figure~\ref{fig:mcr_by_subgenre}.
Since the observation-behavior gap is smaller in this subgenre, clue-based reasoning provides less advantage, highlighting differences across subgenres.

\subsection{Ablation Study}
We conducted an ablation study of COAST using Claude to isolate the contributions of the Seeker, Mapper and Solver components.

The results in Table~\ref{tab:coast_ablation} demonstrate that integrating all three components yields the highest overall performance.
When the Seeker was used alone, performance was notably poor across all five tested games.
This indicates that merely identifying clues without a subsequent planning and solving phase is insufficient.
Furthermore, the Seeker + Solver revealed suboptimal milestone completion rate (MCR), particularly in clue-rich mystery/detective and room escape games.
This highlights the critical role of the Mapper in formulating effective subtask plans by connecting raw observations and clues.

\begin{table}[t!]
\centering
\begin{adjustbox}{width=\columnwidth}
\begin{tabular}{llccc}
\toprule
\textbf{Game (Subgenre)} & \textbf{Metric} & \textbf{Seeker} & \textbf{Seeker + Solver} & \textbf{\methodName} \\
\midrule
\multirow{3}{*}{\makecell[l]{Sherlock Holmes 2 \\(Mystery/Detective)}} 
  & Success (\%)   & 0.0   & 0.0   & 0.0   \\
  & Milestone (\%) & 37.5  & 37.5  & 62.5  \\
  & \# Steps       & 1000  & 1000  & 1000  \\
\midrule
\multirow{3}{*}{\makecell[l]{Grim Tales: The Bride \\(Hidden Object)}} 
  & Success (\%)   & 0.0   & 100.0 & 100.0 \\
  & Milestone (\%) & 66.7  & 100.0 & 100.0 \\
  & \# Steps       & 1000  & 790   & 225   \\
\midrule
\multirow{3}{*}{\makecell[l]{Camping Room Escape \\(Room Escape)}} 
  & Success (\%)   & 0.0   & 0.0   & 0.0   \\
  & Milestone (\%) & 22.2  & 33.3  & 44.4  \\
  & \# Steps       & 1000  & 1000  & 1000  \\
\midrule
\multirow{3}{*}{\makecell[l]{Idol Days Sim Date \\(Visual Novel)}} 
  & Success (\%)   & 0.0   & 0.0   & 0.0   \\
  & Milestone (\%) & 2.3   & 25.6  & 25.6  \\
  & \# Steps       & 1000  & 1000  & 1000  \\
\midrule
\multirow{3}{*}{\makecell[l]{Sort the Court \\(Simulation)}} 
  & Success (\%)   & 0.0   & 0.0   & 0.0   \\
  & Milestone (\%) & 83.2  & 93.0  & 95.5  \\
  & \# Steps       & 1000  & 1000  & 1000  \\
\midrule
\multirow{3}{*}{\textbf{Total}} 
  & Success (\%)   & 0.0   & \textbf{20.0}  & \textbf{20.0}  \\
  & Milestone (\%) & 42.4  & \underline{57.9}  & \textbf{65.6}  \\
  & \# Steps       & 1000  & 958   & 845   \\
\bottomrule
\end{tabular}
\end{adjustbox}
\caption{Comparison across subgenres under different agent configurations, based on five sampled games.}
\label{tab:coast_ablation}
\end{table}

In conclusion, this ablation study confirms that all three modules are indispensable and contribute to COAST's optimal performance, thereby validating the necessity of our proposed architecture.

\subsection{Failure Analysis and Mitigation}

To analyze how \methodName mitigates agent failures, we manually monitored gameplay and categorized errors into (1) weak planning, (2) poor visual perception, (3) deficient lateral thinking, and (4) inefficient resource management (Table~\ref{tab:failure_cases}).

\begin{table}[h!]
\centering
\begin{adjustbox}{width=\columnwidth}
\begin{tabular}{lccc}
\toprule
\textbf{Game (Subgenre)} & 
\makecell{\textbf{GPT-4o} \\ \textbf{+ UGround} \\ + \textbf{Cradle}} & 
\makecell{\textbf{Claude-3.7} \\ \textbf{Computer-Use}} & 
\makecell{\textbf{Claude-3.7} \\ \textbf{Computer-Use} \\ \textbf{+ COAST}} \\
\midrule
\makecell[l]{Sherlock Holmes 2 \\ (Mystery/Detective)} & 1,2,3 & 1,2,3 & 1,2 \\
\midrule
\makecell[l]{Grim Tales: The Bride \\ (Hidden Object)} & 1,2   & 1     & -   \\
\midrule
\makecell[l]{Camping Room Escape \\ (Room Escape)}     & 1,2,3 & 1,2,3 & 1,2 \\
\midrule
\makecell[l]{Idol Days Sim Date \\ (Visual Novel)}     & 4     & 4     & 4   \\
\midrule
\makecell[l]{Sort the Court \\ (Simulation)}           & 4     & 4     & -   \\
\bottomrule
\end{tabular}
\end{adjustbox}
\caption{Comparison of failure patterns across agents, with codes denoting failure types: 1 = planning, 2 = perception, 3 = lateral thinking, 4 = resource management.}
\label{tab:failure_cases}
\end{table}

\methodName improved \textbf{planning} in Grim Tales (hidden object), where it successfully used prior failed attempts to locate hidden objects.
It also alleviated \textbf{lateral thinking} issues in Sherlock Holmes 2 (mystery/detective) and Camping Room Escape (room escape), where it bridged long-term observation–behavior gaps (\eg using a blue ball to retrieve a card on the ceiling).

In contrast, results were mixed for \textbf{resource management}: while \methodName made more efficient yes/no choices in Sort the Court (simulation), it failed to gain high experience score in Idol Days Sim Date (visual novel).
Finally, \methodName did not mitigate \textbf{visual perception} failures, as it inherits the same backbone (Claude-Sonnet-3.7). GPT-4o + UGround + Cradle was especially prone to such perception errors in Grim Tales, likely due to UGround’s limited GUI grounding.

Overall, these findings highlight our method’s strength in mitigating planning and lateral thinking failures, while limitations remain in perception and resource management.

\section{Further Analysis}

\subsection{Analysis on Contaminations}
We investigate whether LLMs were exposed to game-specific information during pretraining and how such contamination affected their in-game performance~\citep{paglieri2025balrog}. 
We analyze Claude-3.7-Sonnet and GPT-4o.
For each subgenre, we selected a representative game and constructed questions targeting specific in-game situations (\eg actions, interactions, problem-solving steps). 
These questions assessed whether the models possessed pretrained knowledge of the scenarios and could answer correctly without genuine reasoning.

Claude-3.7-Sonnet showed no signs of contamination, suggesting fair problem solving without memorized content. 
In contrast, GPT-4o exhibited contamination and hallucination in 3 of 10 questions. 
In one case, it revealed knowledge of game-specific content but failed to execute the corresponding action in gameplay.
This highlights a \textit{knowing-doing gap}~\citep{paglieri2025balrog}: despite contaminated knowledge, GPT-4o’s gameplay performance (Table~\ref{tab:performance_comparison_full}) lagged behind Claude-3.7-Sonnet, showing its inability to translate knowledge into action. 
See Appendix~\ref{subsec:appendix_contamination} for details.

\subsection{Additional Hint Injection}
We evaluate Claude-3.7-Sonnet Computer-Use with and without injected hints (\ie oracle subtasks), representing an upper performance bound.

In \textit{Sherlock Holmes: The Tea Shop Murder Mystery}, hints enabled the agent to complete all milestones in 758 turns.
Without them, it only achieved one milestone after 1,000 turns. 
This shows that explicit subtasks are highly effective for navigation and problem-solving in well-structured environments.
In contrast, for \textit{Computer Office Escape}, which demands complex pattern recognition, the agent failed to clear the first milestone even with hints.
This suggests that while hints can streamline task execution, they do not overcome fundamental bottlenecks in complex spatial-logical reasoning over patterns in unstructured environments.
Further details and results using a Large Reasoning Model (o4-mini) are available in Appendix~\ref{subsec:appendix_additional_hint_injection} and Appendix~\ref{subsec:large_reasoning_models}.

\section{Conclusion}
We presented \benchmarkName, a benchmark of \numGames diverse Flash-based adventure games designed to evaluate GUI agents’ ability to complete full story arcs.
To enable reliable evaluation, we introduced CUA-as-a-Judge, an automatic milestone verification agent that closely aligns with human judgments.
Our experiments showed that current state-of-the-art GUI agents struggle with planning, perception, and lateral thinking required for full game completion. Therefore, we proposed \methodName, a clue-oriented framework that manages long-term clue memory and generates subtasks, significantly improving performance and bridging the observation-behavior gap.
Despite these advances, a substantial gap remains between agent and human performance, requiring further improvements.

\section*{Limitations}
A limitation lies in the reliance on manually defined milestones, which is common across prior video game benchmarks~\citep{chen2024varp,tan2024cradle,zhang2025videogamebench}. 
To mitigate concerns about \textit{labor intensity}, we designed a lightweight process requiring only 10-15 minutes per game, making it unlikely to become a bottleneck when scaling. 
For each game, (1) an annotator first selected initial milestones (\eg clues, items, events) while reviewing the full walkthrough (3-5 minutes), and (2) evaluated whether each was indispensable for completing the story, discarding the rest (7-10 minutes). 
To enhance the \textit{objectivity} of milestone selection, future work could leverage narrative flow charts~\citep{paschali2018tool} to derive milestones directly from structured story representations, providing a more principled evaluation framework.

In addition, our evaluation method, CUA-as-a-Judge, is not fully genre-agnostic.
It is suitable within \benchmarkName because selected games lack strict time constraints and narrative progression does not directly depend on character movements (\S~\ref{subsec:game_selection}).
However, this approach is less applicable to fast-paced action games (\eg, \textit{Super Mario Bros}), where character movement directly impacts the game state or narrative outcome.
Exploring extensions to such reflex-oriented genres, such as pausing gameplay to report stats via clicks, would be an interesting direction for future work.

As mentioned in \S~\ref{sec:approach}, we didn't implement any memory management for \methodName, since each clue is small enough and the total input didn't exceed the maximum memory limit.
However, this may lead to scalability issues as the number of action steps increases.
To address this, one can potentially apply summarization~\citep{tan2024cradle}, retrieval~\citep{agashe2024agents}, or forgetting~\citep{park2023generativeagents} mechanisms to effectively manage memory.

Lastly, running all games for 1,000 steps using proprietary APIs (\eg GPT-4o, Claude-3.7-Sonnet) incurs substantial cost, posing a barrier to reproduction and large-scale experimentation.
However, this issue is not inherent to our benchmark itself, but rather reflects the current need for more efficient open-source GUI agents to reduce reliance on expensive API calls.

\section*{Ethics Statement}
Regarding game copyright, all games included in our benchmark remain the intellectual property of their original creators and publishers.
We emphasize that our use of these games is strictly for \textbf{non-commercial, academic research purposes} only.
Before including the games, we contacted the respective rights holders (such as developers and publishers) to obtain permission and clarify our intended use.
Our usage is limited to evaluation rather than model training, and we do not redistribute or modify the original software.
Instead, the games are accessed via authorized platforms like Flashpoint Archive, preserving the original experience.
We rely solely on screenshot-based observations for the gameplay, without reverse-engineering or modding any low-level game code.
This approach minimizes potential copyright infringement and respects the integrity of the original works.

For the collection of human gameplay demonstrations, we recruited 13 adult participants who provided informed consent prior to the study.
Participants were compensated fairly at or above the national minimum wage for each game played, regardless of completion.
All procedures, including data collection via input logging and screen recording, were conducted with respect for participant privacy and autonomy.
Participants were free to withdraw at any time without penalty. 
We designed this process to minimize any risks, such as fatigue, and ensure ethical treatment throughout the gameplay sessions.
Please refer to Appendix~\ref{subsec:appendix_human_game_play} for further details.

\section*{Acknowledgments}
We thank all annotators for their dedication and invaluable contributions to the development of \benchmarkName.
We are also grateful to Euihyun Tae, Junseo Koo, Wonkwang Lee, Sangwoo Moon, Seyeon Choi, Sieon Park, Kangwook Lee, and the anonymous reviewers for their insightful feedback.
This work was supported by the Institute of Information \& Communications Technology Planning \& Evaluation (IITP) grant funded by the Korea government (MSIT) (No.~RS-2019-II191082, RS-2021-II211343, No.~RS-2022-II220156), 
the Basic Science Research Program through the National Research Foundation of Korea (NRF) funded by the Ministry of Education (RS-2023-00274280), 
and the Korea Radio Promotion Association (Development of Intelligent Docent Service for Information-Disadvantaged Groups). 
Gunhee Kim is the corresponding author.


\clearpage
\appendix

\section{Details on \benchmarkName}
\label{sec:appendix_benchmark_details}

\subsection{Related Work}
\label{subsec:appendix_related_work}
\textbf{Text Adventure Games.}
Research on text-based adventure games has long served as a testbed for studying reasoning, planning, and long-horizon decision-making in language agents.
Classic environments such as TextWorld~\citep{cote2018textworld} and ScienceWorld~\citep{wang2022scienceworld} provide text-based interfaces where agents must parse observations, generate natural language actions, and interact with simulated environments.
Jericho~\citep{hausknecht2020jericho} is perhaps the most relevant to our setting, as games in the Jericho suite (\eg Zork) involve long-term dependencies, rich narrative arcs, and extremely large action spaces, posing challenges similar to those in open-ended video games.
More recent benchmarks such as Jiminy Cricket~\citep{hendrycks2021jiminy} also extend this paradigm, focusing on narrative coherence and ethical reasoning.

Our proposed \benchmarkName can be viewed as bridging this line of text-adventure research with the broader video game domain.
While traditional text-adventure environments emphasize symbolic reasoning and narrative progression, \benchmarkName introduces multimodal, interactive settings that preserve the long-horizon and combinatorial complexity of Jericho-like games while grounding them in video game worlds.
This connection highlights the role of \benchmarkName as a unifying benchmark, linking interactive fiction with modern video game environments for evaluating LLM-based GUI agents.

\subsection{Problem Formulation}
\label{subsec:appendix_problem_formulation}

Table~\ref{tab:action_space} presents the complete action set $\mathcal{A}$ used by GUI agents.

\label{subsec:problem_formulation}
\begin{table}[h!]
\centering
\begin{adjustbox}{width=\columnwidth}
\begin{tabular}{ll}
\toprule
\textbf{Action} & \textbf{Description} \\
\midrule
\texttt{left\_click(x, y)}      & Left-click at coordinates. \\
\texttt{right\_click(x, y)}     & Right-click at coordinates. \\
\texttt{middle\_click(x, y)}    & Middle-click at coordinates. \\
\texttt{double\_click(x, y)}    & Double-click at coordinates. \\
\texttt{triple\_click(x, y)}    & Triple-click at coordinates. \\
\texttt{drag\_from\_(x1,y1)\_to\_(x2,y2)} & Drag with mouse to coordinates. \\
\texttt{scroll(dir, amount)}    & Scroll screen (up, down, left, right). \\
\midrule
\texttt{key: enter}             & Press a specific key (\eg, Enter). \\
\texttt{type text: hello}       & Type text ``hello''. \\
\texttt{hold\_key(ctrl, 2)}     & Hold key for duration (\eg, 2s). \\
\midrule
\texttt{finish}     & The task is finished. \\
\bottomrule
\end{tabular}
\end{adjustbox}
\caption{Unified action space $\mathcal{A}$.}
\label{tab:action_space}
\end{table}

\subsection{Game Selection}
\label{subsec:appendix_game_selection}
We provide a complete list of games available in \benchmarkName~organized by subgenre:

\begin{table}[h!]
\centering
\small
\begin{adjustbox}{width=\columnwidth}
\begin{tabular}{ll}
\toprule
\textbf{Subgenre} & \textbf{Games} \\ 
\midrule
\makecell[l]{Point-and-Click Adventure\\(Mystery/Detective)} &
\begin{tabular}[c]{@{}l@{}}
Sherlock Holmes: The Tea Shop Murder Mystery \\
Sherlock Holmes 2 \\
Vortex Point 1 \\
Vortex Point 2 \\
Vortex Point 3 \\
Pierre Hotel \\
Small Town Detective \\
Dakota Winchester’s Adventures \\
Saucy Devil Gordon \\
Ray and Cooper 2 \\
Nick Bounty: A Case of the Crabs
\end{tabular} \\ 
\midrule
Hidden Object & 
\begin{tabular}[c]{@{}l@{}}
Grim Tales: The Bride \\ 
Grim Tales: The Legacy Collector’s Edition
\end{tabular} \\ 
\midrule
Room Escape & 
\begin{tabular}[c]{@{}l@{}}
Computer Office Escape \\ 
Crimson Room \\ 
Camping Room Escape \\ 
Chemical Room Escape \\ 
Space Museum Escape \\ 
Vending Machine Room \\ 
Wood Workshop Escape \\ 
Geometric Room Escape \\ 
Game Cafe Escape \\ 
Machine Room Escape \\ 
VideoStudio Escape \\ 
Design House Escape \\ 
Paint Room Escape \\ 
Mirror Room Escape \\ 
Elevator Room Escape
\end{tabular} \\ 
\midrule
Visual Novel (Dating Sim) & 
\begin{tabular}[c]{@{}l@{}}
Pico Sim Date \\ 
Festival Days Sim Date \\ 
Kingdom Days \\ 
Idol Days Sim Date
\end{tabular} \\ 
\midrule
Simulation: Life & 
\begin{tabular}[c]{@{}l@{}}
Community College Sim \\ 
\end{tabular} \\
\midrule
Simulation: Management & Sort the Court \\ 
\bottomrule
\end{tabular}
\end{adjustbox}
\caption{List of games included in \benchmarkName.}
\label{table:game_selection}
\end{table}

\begin{figure*}[h!]
    \centering
    \includegraphics[width=\linewidth]{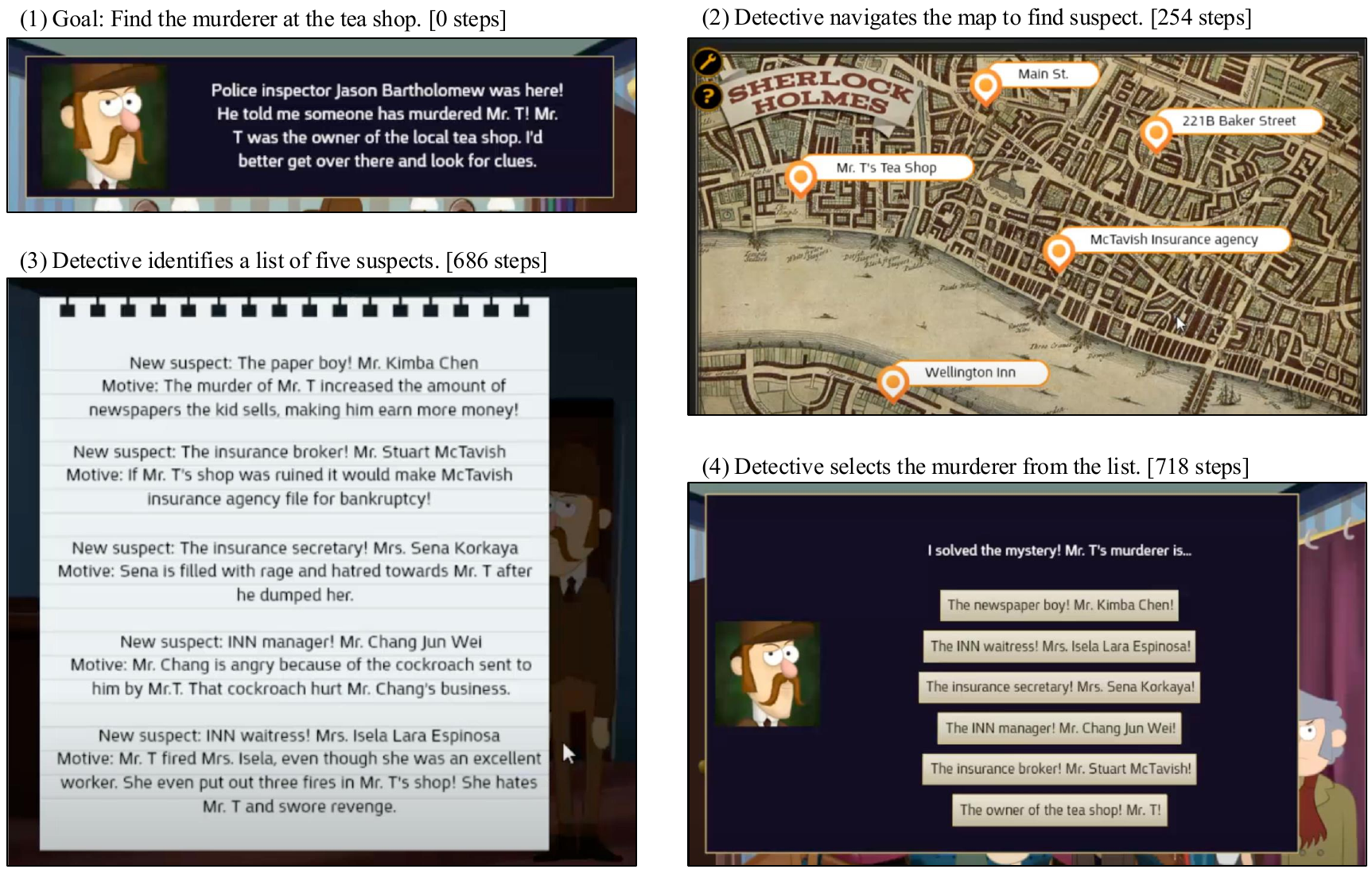}
    \caption{An illustration of the \textit{Point-and-Click Adventure (mystery/detective)} subgenre, showing a human player's walkthrough of \textit{Sherlock Holmes: The Tea Shop Murder Mystery}. The cumulative number of steps is written at the end of each subfigure caption; the game ends at 718 steps.}
    \label{fig:walkthrough_sherlock}
\end{figure*}

\begin{figure*}[h!]
    \centering
    \includegraphics[width=\linewidth]{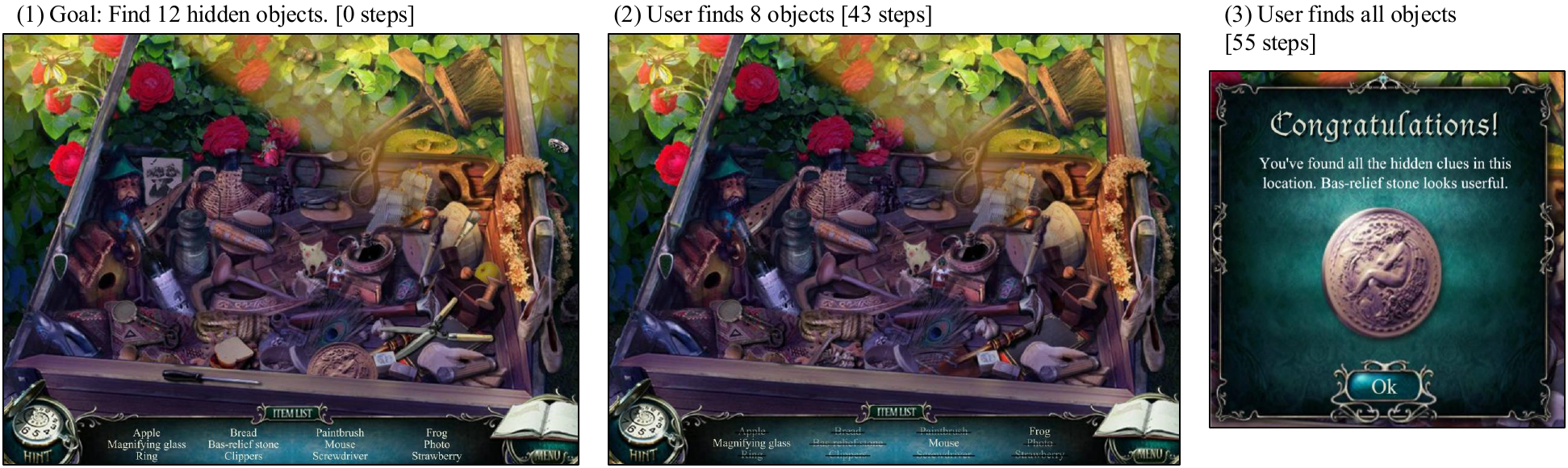}
    \caption{An illustration of the \textit{Hidden Object} subgenre, showing a human player's walkthrough of \textit{Grim Tales: The Bride}. The cumulative number of steps is written at the end of each subfigure caption; the game ends at 55 steps.}
    \label{fig:walkthrough_grim_tales}
\end{figure*}

\begin{figure*}[h!]
    \centering
    \includegraphics[width=\linewidth]{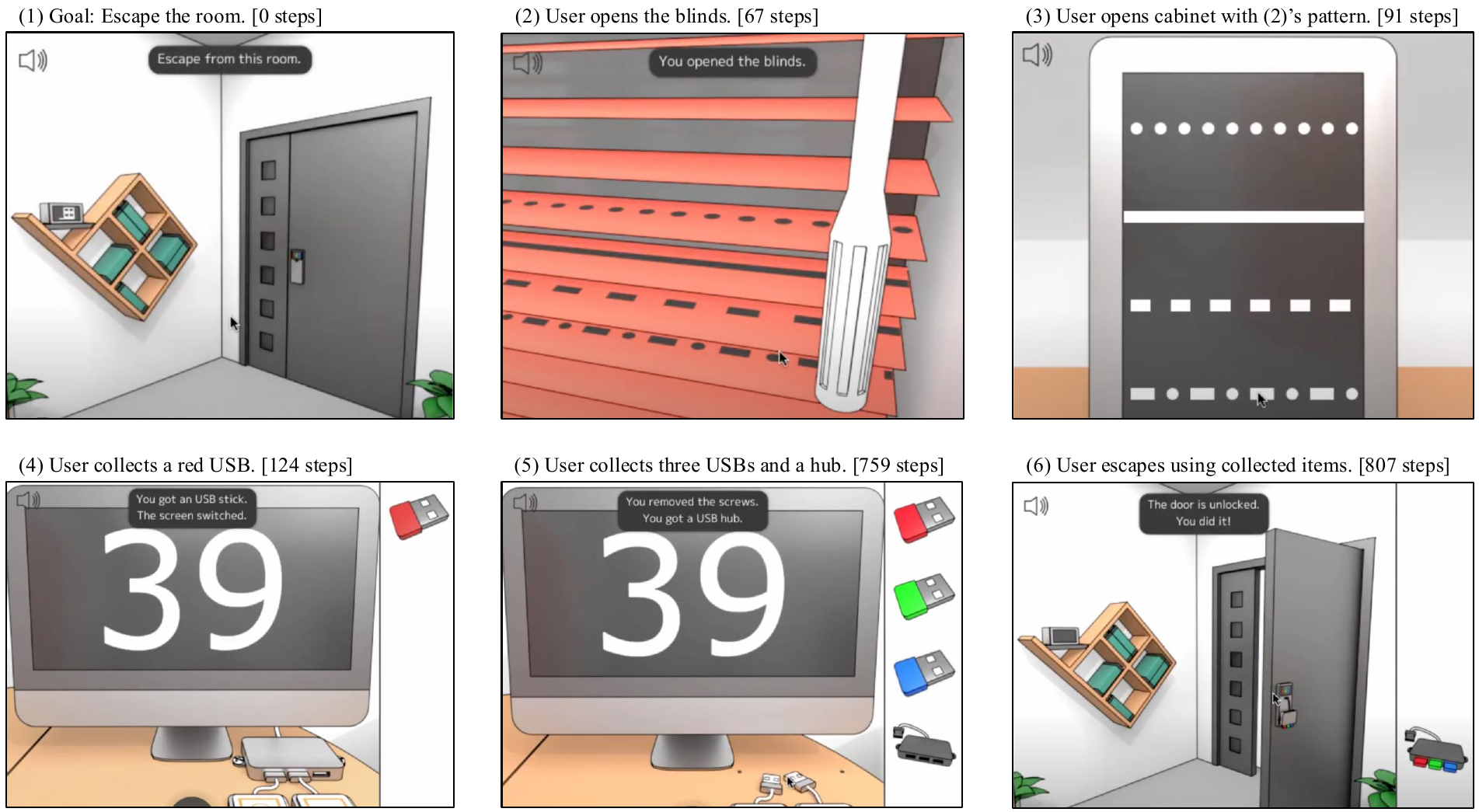}
    \caption{An illustration of the \textit{Room Escape} subgenre, showing a human player's walkthrough of \textit{Computer Office Escape}. The cumulative number of steps is written at the end of each subfigure caption; the game ends at 807 steps.}
    \label{fig:walkthrough_computer_room_escape}
\end{figure*}

\begin{figure*}[h!]
    \centering
    \includegraphics[width=\linewidth]{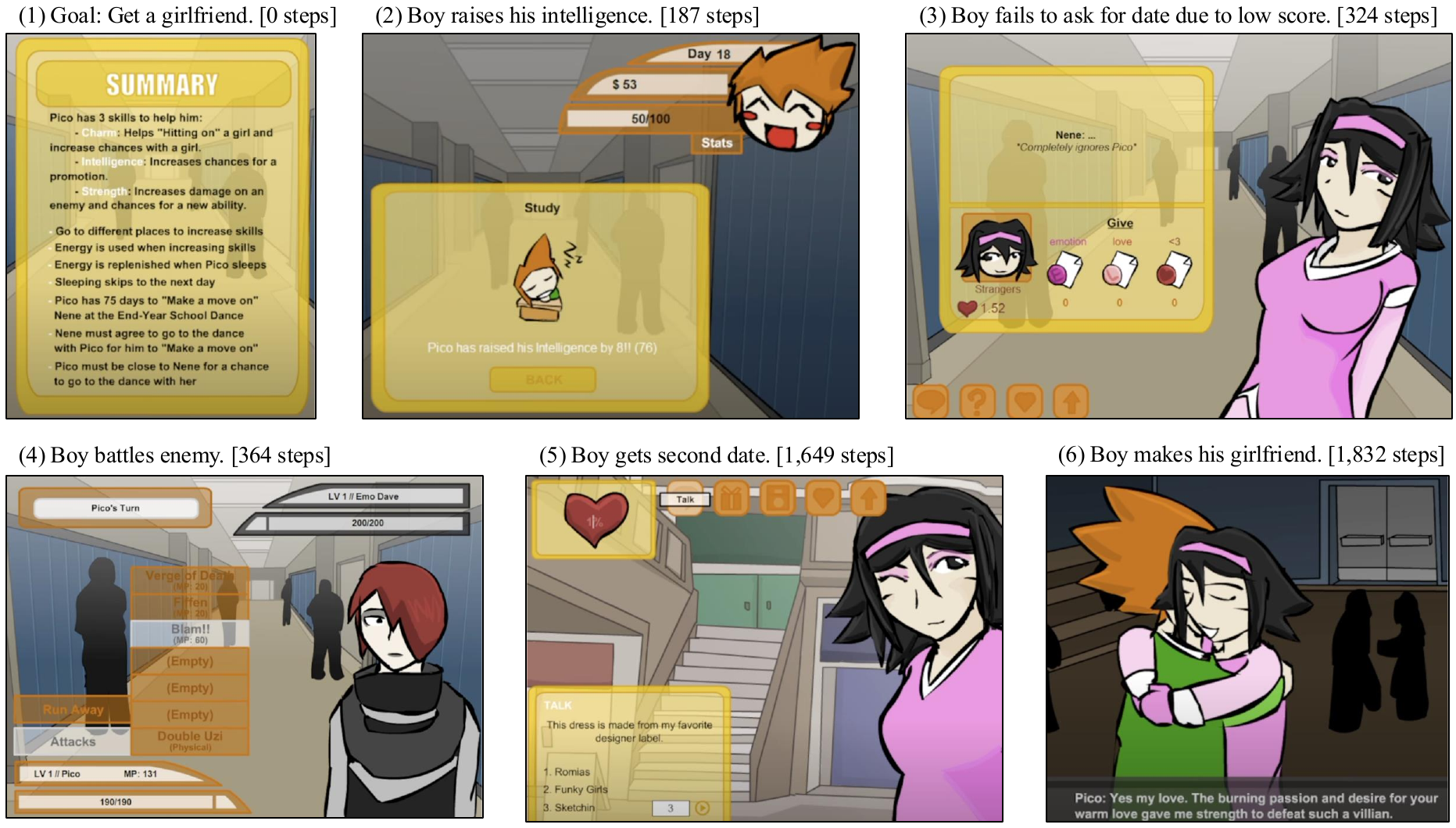}
    \caption{An illustration of the \textit{Visual Novel (Dating Sim)} subgenre, showing a human player's walkthrough of \textit{Pico Sim Date}. The cumulative number of steps is written at the end of each subfigure caption; the game ends at 1,832 steps.}
    \label{fig:walkthrough_pico_sim}
\end{figure*}

\begin{figure*}[h!]
    \centering
    \includegraphics[width=\linewidth]{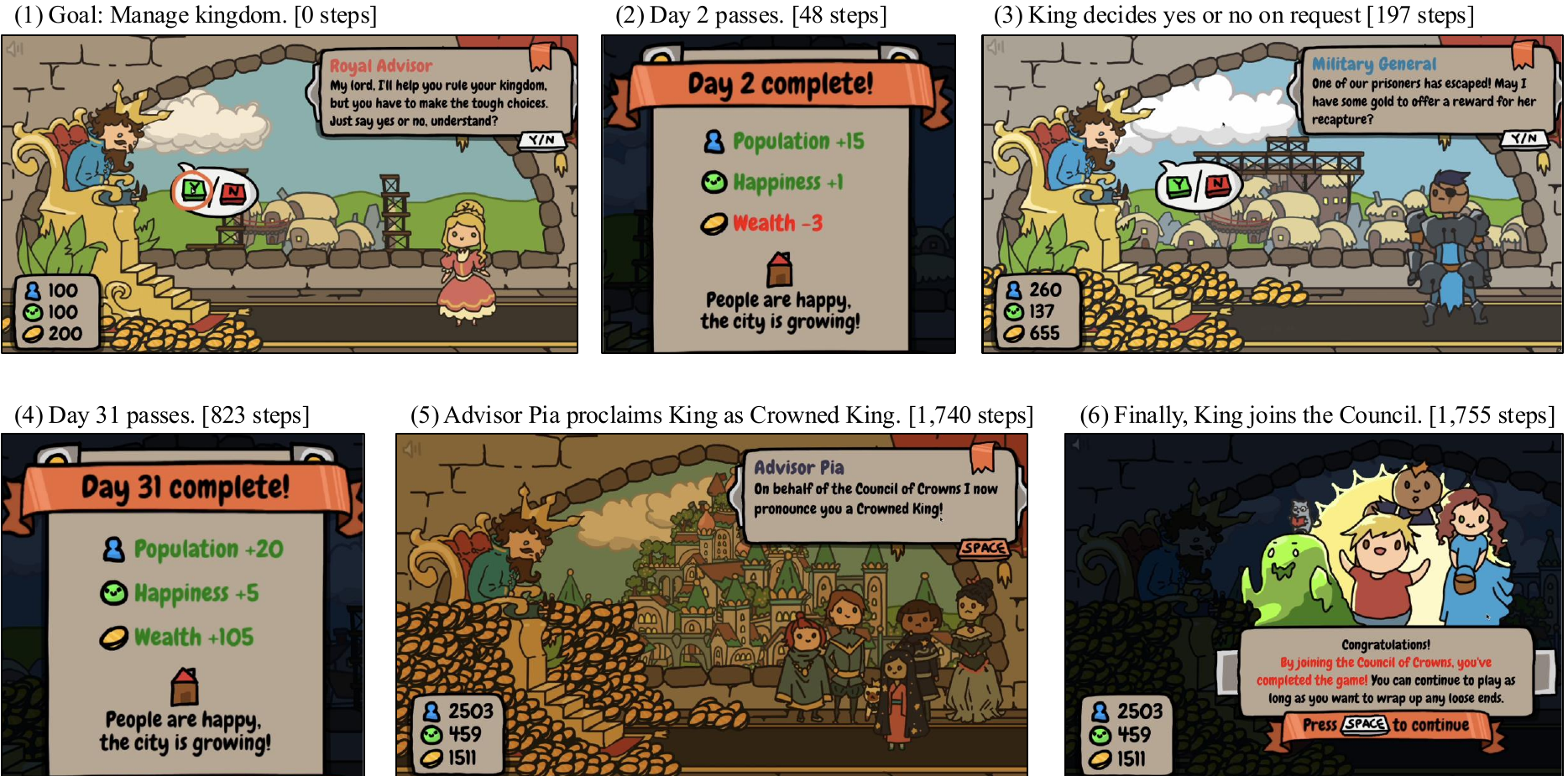}
    \caption{An illustration of the \textit{Simulation} subgenre, showing a human player's walkthrough of \textit{Sort the Court}. The cumulative number of steps is written at the end of each subfigure caption; the game ends at 1,755 steps.}
    \label{fig:walkthrough_sort_the_court}
\end{figure*}

In addition, we provide human walkthroughs for one selected game per subgenre, including annotated screenshots and step counts.
Specifically, Figure~\ref{fig:walkthrough_sherlock} presents a \textit{point-and-click adventure (mystery/detective)} example (\textit{Sherlock Holmes: The Tea Shop Murder Mystery}), Figure~\ref{fig:walkthrough_grim_tales} shows a \textit{hidden object} example (\textit{Grim Tales: The Bride}), Figure~\ref{fig:walkthrough_computer_room_escape} depicts a \textit{room escape example} (\textit{Computer Office Escape}), Figure~\ref{fig:walkthrough_pico_sim} provides a \textit{visual novel (dating sim)} example (\textit{Pico Sim Date}), and Figure~\ref{fig:walkthrough_sort_the_court} illustrates a \textit{simulation} example (\textit{Sort the Court}).

\begin{table*}[h!]
\centering
\begin{adjustbox}{width=\linewidth}
\begin{tabular}{lccccc}
\toprule
\textbf{Subgenre} & \textbf{Perceptual Precision} & \textbf{Logical Reasoning} & \textbf{Lateral Thinking} & \textbf{Social Reasoning (Conversation with NPC)} & \textbf{Resource Management} \\
\midrule
Mystery/Detective & High & High & High & High & Low \\
Hidden Object     & High & Low  & Mid  & Low  & Low \\
Room Escape       & High & High & High & Low  & Low \\
Visual Novel      & Mid  & Mid  & Low  & High & High \\
Simulation        & Mid  & High & Low  & High & High \\
\bottomrule
\end{tabular}
\end{adjustbox}
\caption{Inter-subgenre diversity: comparison of subgenres across cognitive dimensions.}
\label{tab:inter_subgenre_diversity}
\end{table*}

\begin{table*}[h!]
\centering
\begin{adjustbox}{width=\linewidth}
\begin{tabular}{lrrrrrrlrrr}
\toprule
\textbf{Games} & \textbf{Steps (A)} & \textbf{Steps (B)} & \textbf{Steps (C)} & \textbf{Rank (A)} & \textbf{Rank (B)} & \textbf{Rank (C)} & \textbf{Best Player} & \textbf{Norm. Steps (A)} & \textbf{Norm. Steps (B)} & \textbf{Norm. Steps (C)} \\
\midrule
\multicolumn{11}{l}{\textbf{Mystery/Detective}} \\
Sherlock Holmes: The Tea Shop Murder Mystery & 718  & 439  & 752  & 2 & 1 & 3 & B & 81.7  & -197.3 & 115.7 \\
Sherlock Holmes 2                             & 399  & 1037 & 823  & 1 & 3 & 2 & A & -354.0 & 284.0  & 70.0  \\
Vortex Point 1                                & 707  & 1461 & 794  & 1 & 3 & 2 & A & -280.3 & 473.7  & -193.3 \\
Vortex Point 2                                & 1147 & 595  & 646  & 3 & 1 & 2 & B & 351.0  & -201.0 & -150.0 \\
Vortex Point 3                                & 466  & 588  & 503  & 1 & 3 & 2 & A & -53.0  & 69.0   & -16.0  \\
Pierre Hotel                                  & 862  & 318  & 1013 & 2 & 1 & 3 & B & 131.0  & -413.0 & 282.0  \\
Small Town Detective                          & 437  & 1103 & 905  & 1 & 3 & 2 & A & -378.0 & 288.0  & 90.0   \\
Dakota Winchester's Adventures                & 326  & 568  & 502  & 1 & 3 & 2 & A & -139.3 & 102.7  & 36.7   \\
Saucy Devil Gordon                            & 297  & 387  & 455  & 1 & 2 & 3 & A & -92.7  & 7.3    & 75.3   \\
Nick Bounty: A Case of the Crabs              & 1292 & 1660 & 531  & 2 & 3 & 1 & C & 131.0  & 499.0  & -630.0 \\
\midrule
\multicolumn{11}{l}{\textbf{Room Escape}} \\
Computer Office Escape & 1625 & 747 & 741 & 3 & 2 & 1 & C & 587.3 & -290.7 & -296.7 \\
Camping Room Escape & 1430 & 2376 & 1141 & 2 & 3 & 1 & C & -219.0 & 727.0 & -508.0 \\
Space Museum Escape & 1283 & 1119 & 1936 & 2 & 1 & 3 & B & -163.0 & -327.0 & 490.0 \\
Wood Workshop Escape & 848 & 1452 & 947 & 1 & 3 & 2 & A & -234.3 & 369.7 & -135.3 \\
Geometric Room Escape & 1247 & 2221 & 1099 & 2 & 3 & 1 & C & -275.3 & 698.7 & -423.3 \\
Game Cafe Escape & 1280 & 1091 & 1262 & 3 & 1 & 2 & B & 69.0 & -120.0 & 51.0 \\
Machine Room Escape & 896 & 903 & 1206 & 1 & 2 & 3 & A & -105.7 & -98.7 & 204.3 \\
VideoStudio Escape & 1513 & 2341 & 2330 & 1 & 3 & 2 & A & -548.3 & 279.7 & 268.7 \\
Design House Escape & 1671 & 1402 & 1482 & 3 & 1 & 2 & C & 152.7 & -116.3 & -36.3 \\
Elevator Room Escape & 1227 & 1194 & 2362 & 2 & 1 & 3 & B & -367.3 & -400.3 & 767.7 \\
\bottomrule
\end{tabular}
\end{adjustbox}
\caption{Intra-subgenre diversity: Performance of human participants (A, B, C) in mystery/detective and room escape games, measured by steps, ranks, and normalized steps (\ie \# steps - \# average steps per game).}
\label{tab:intra_subgenre_diversity}
\end{table*}

\subsection{Analysis on Game Diversity}
\label{subsec:analysis_game_diversity}

To clarify that our benchmark offers substantial diversity and presents a challenging testbed for modern GUI agents, we explain the diversity of \benchmarkName along two dimensions: \textit{inter-subgenre} and \textit{intra-subgenre diversity}.

\textbf{Inter-subgenre diversity}: \benchmarkName includes five subgenres, each of which is thematically and structurally distinct, as shown in the Table~\ref{tab:inter_subgenre_diversity}.
Mystery/detective games require social interactions with NPCs to gather key clues (\eg characters explore towns and engage with various people), and room escape games focus on solving complex puzzles in isolated environments.
The hidden object subgenre is unique in that it heavily relies on precise visual perception to identify hidden items within a scene. 
These games also tend to have the shortest playtime (less than 5 minutes) among all subgenres.
Finally, visual novel and simulation games both emphasize efficient resource management to achieve high scores,
However, they differ in their core objectives: visual novels typically focus on optimizing a single key metric, such as an affection with a specific character, while simulation games often require balancing multiple factors (\eg wealth, population, happiness), demanding a higher degree of logical reasoning.

\textbf{Intra-subgenre diversity}: Among our 34 games, a fraction display thematically similar narratives, \eg mystery/detective games.
Nevertheless, seemingly relevant games display considerably diverse structures; excelling in one game does not necessarily lead to excelling in another.
We provide a game-play statistics collected from three human players who solved 10 mystery/detective games and 10 room escape games in common, as shown in Table~\ref{tab:intra_subgenre_diversity}.
The results indicate that each player tends to excel in different games. 

\begin{itemize}
\item Across both subgenres, rankings and normalized step scores varied substantially across games. 
These fluctuations were reflected in the sign and magnitude of each player’s score, which shifted significantly from game to game, indicating high performance variability. 

\item \textbf{Mystery/Detective:} Player A outperformed the others most frequently, but the standard deviations of normalized scores were still large (A=236, B=302, C=249).

\item \textbf{Room Escape:} Performance variability was even greater, with standard deviations of normalized scores (A=314, B=361, C=964).
\end{itemize}

To sum it up, \benchmarkName exhibits both inter- and intra-subgenre diversity, as evidenced by distinct gameplay requirements across subgenres and substantial performance variability among human players within the same subgenre.

\subsection{Evaluation}
\label{subsec:appendix_evaluation}
\noindent\textbf{Full List of Milestones and Game Introductions.}
In this section, we provide the full list of milestones for each game, along with a brief introduction to each game.

{\small
\begin{enumerate}
\item \textbf{Sherlock Holmes: The Tea Shop Murder Mystery (6 milestones)}: Detective/Mystery game where a player investigates the accidental death of a tea shop owner. \underline{Milestone}: (1) Verify that the player checks the number of new suspects via the notebook item (5 total), (2) Verify correct suspect selection in the final ending.
\item \textbf{Sherlock Holmes 2 (8 milestones)}: Detective/Mystery game where a player investigates a murder at a mansion. \underline{Milestone}: (1) Check collection of essential items (5 total), (2) Confirm that the fire alarm system on the second floor of the crime scene is opened, (3) Confirm whether the outdoor area of the crime scene has been set on fire, (4) Verify task completion through to the final ending.
\item \textbf{Vortex Point 1 (7 milestones)}: Detective/Mystery game where a player solves a supernatural theft in the town of Vortex Point. \underline{Milestone}:
(1) Verify the discovery of 4 hidden locations in the given map,  
(2) Confirm successful entry into 2956 Vineyard Drive,  
(3) Confirm successful entry into C. Razy Mental Hospital,  
(4) Verify task completion through to the final ending.

\item \textbf{Vortex Point 2 (5 milestones)}: Detective/Mystery game where a player investigates the paranormal disappearance of a girl linked to a mysterious photo booth.  \underline{Milestone}:
(1) Check if the man inside the hidden pub at the crime scene has disappeared,  
(2) Check if the magician inside the yellow-sign building on Main Street has disappeared and if water is spilling,  
(3) Verify disappearance of the guard at Town’s History Museum,  
(4) Confirm that the fence at Marshall Square has been opened,  
(5) Verify task completion through to the final ending.

\item \textbf{Vortex Point 3 (5 milestones)}: Detective/Mystery game where a player investigates the disappearance of a tourist at Vortex Lake Hotel. \underline{Milestone}:
(1) Check if the man at Vortex Lake Hotel is unexpectedly eating a hamburger,  
(2) Confirm whether the toilet in the Souvenir Shop is accessible,  
(3) Verify the disappearance of the man in the Souvenir Shop,  
(4) Check if the player can board the boat at Vortex Lake Pier (\ie reach Private Property),  
(5) Verify task completion through to the final ending.

\item \textbf{Pierre Hotel (6 milestones)}: Detective/Mystery game where a player searches for their missing girlfriend in a hotel full of vampires. \underline{Milestone}:
(1) Verify that the wine bar staff on the first floor is on a phone call,  
(2) Confirm access to the broom room on the second floor,  
(3) Check if the fireplace on the first floor has been extinguished,  
(4) Confirm that the elevator on the first floor operates properly,  
(5) Verify disappearance of the vampire at the first floor (front counter),  
(6) Verify task completion through to the final ending.

\item \textbf{Small Town Detective (6 milestones)}: Detective/Mystery game where a player helps Paul the Detective find a missing journalist in a small town. \underline{Milestone}:
(1) Verify the discovery of 5 hidden locations (excluding Paul’s Office) in the given map,  
(2) Verify task completion through to the final ending.

\item \textbf{Dakota Winchester’s Adventures (6 milestones)}: Detective/Mystery game where a player guides archaeologist Dakota Winchester on a quest to find cursed rubies and unlock Hilda’s box. \underline{Milestone}:
(1) Confirm that the stepping stones to the bomb shelter have been placed,  
(2) Check if a fire is lit in the inner village area,  
(3) Verify that the temple entrance is open,  
(4) Confirm that a monkey is eating a banana inside the temple,  
(5) Check if a yellow ceiling light is illuminating the temple,  
(6) Verify task completion through to the final ending.

\item \textbf{Saucy Devil Gordon (5 milestones)}: Detective/Mystery game where a player helps Gordon become a pirate. \underline{Milestone}:
(1) Confirm that a coconut has been retrieved from a palm tree,  
(2) Check whether a pineapple has been picked next to the door of the left house in the scene with a tiki statue,  
(3) Verify that the door in the skull-marked area is open,  
(4) Confirm that light is shining vertically on the grave,  
(5) Verify task completion through to the final ending.

\item \textbf{Ray and Cooper 2 (6 milestones)}: Detective/Mystery game where a player controls investigator Paul Maxstrong to find the missing Ray and Cooper. \underline{Milestone}:
(1) Check whether the upper window frame inside Cooper’s apartment is open,  
(2) Confirm that “MATTEO’S PIZZA” is closed,  
(3) Verify that the door under the “NEW WING! COMING SOON!” banner inside the Museum of Oddities is accessible,  
(4) Confirm the museum manager of the Museum of Oddities has disappeared, and a fence has appeared,  
(5) Check whether the woman on the box in Snack Shop 1162 has disappeared,  
(6) Verify task completion through to the final ending.

\item \textbf{Nick Bounty: A Case of the Crabs (5 milestones)}: Detective/Mystery game where a player helps detective Nick Bounty solve the murder of a seafood salesman. \underline{Milestone}
(1) Verify the discovery of 4 hidden locations in the given map, 
(2) Verify task completion through to the final ending.

\item \textbf{Grim Tales: The Bride (12 milestones)}: Puzzle (hidden object) adventure game where a player finds hidden objects. \underline{Milestone}: 
Verify how many of the 12 hidden objects the player has found in the initial Toolbox scene.

\item \textbf{Grim Tales: The Legacy Collector’s Edition (12 milestones)}: Puzzle (hidden object) adventure game where a player finds hidden objects. \underline{Milestone}:
Verify how many of the 12 hidden objects the player has found in the initial Barrels on the Road scene.

\item \textbf{Computer Office Escape (5 milestones)}: Room escape game where a player solves puzzles and uses items to escape from an office with computers. \underline{Milestone}: 
(1) Verify the collection of three colored USBs (red, green, blue) and a USB hub,  
(2) Verify task completion through to the final ending.

\item \textbf{Crimson Room (14 milestones)}: Room escape game where a player wakes up in a locked crimson-colored room and must find a way out. \underline{Milestone}: 
(1) Verify the collection of all 13 required items,  
(2) Verify task completion through to the final ending.

\item \textbf{Camping Room Escape (9 milestones)}: Room escape game where a player solves puzzles and use items to escape from a room with a tent. \underline{Milestone}:
(1) Verify the collection of all 8 required items,
(2) Verify task completion through to the final ending.

\item \textbf{Chemical Room Escape (8 milestones)}: Room escape game where a player solves puzzles and uses items to escape from a laboratory filled with chemical equipment. \underline{Milestone}:
(1) Verify the collection of all 7 required items,
(2) Verify task completion through to the final ending.

\item \textbf{Space Museum Escape (6 milestones)}: Room escape game where a player is trapped in a space-themed museum and must solve puzzles from exhibits to escape. \underline{Milestone}:
(1–5) Check success based on color-based sequence (red = 0, yellow = 1, green = 2, light blue = 3, blue = 4, pink = 5),
(6) Verify task completion through to the final ending.

\item \textbf{Vending Machine Room (9 milestones)}: Room escape game where a player must use and manipulate vending machines and related items to escape. \underline{Milestone}:
(1) Verify the collection of all 8 required items,
(2) Verify task completion through to the final ending.

\item \textbf{Wood Workshop Escape (7 milestones)}: Room escape game where a player solves puzzles using woodworking tools to escape from a workshop. \underline{Milestone}:
(1) Verify the collection of all 6 required items,
(2) Verify task completion through to the final ending.

\item \textbf{Geometric Room Escape (6 milestones)}: Room escape game where a player must solve puzzles based on geometric patterns to escape. \underline{Milestone}:
(1) Verify the collection of all 5 required items,
(2) Verify task completion through to the final ending.

\item \textbf{Game Cafe Escape (5 milestones)}: Room escape game where a player solves puzzles to escape from a game-themed café. \underline{Milestone}:
(1) Check if the game console has been acquired,  
(2) Verify whether the red box was opened in the mini game,  
(3) Confirm that the player used the gemstone to solve the puzzle and turned off the lights in the mini game,  
(4) Check if the door key was obtained and used in the mini game,
(5) Verify task completion through to the final ending.

\item \textbf{Machine Room Escape (4 milestones)}: Room escape game where a player solves mechanical puzzles to escape from a room filled with machines. \underline{Milestone}:
(1–3) Assess the player's progress in replicating the three distinct door pattern based on tablet illustrations,
(4) Verify task completion through to the final ending.

\item \textbf{VideoStudio Escape (4 milestones)}: Room escape game where a player must solve complex puzzles to escape from a video studio. \underline{Milestone}: 
(1–3) Check how many green lights are lit above the door’s 1, 2, and 3 markers,  
(4) Verify task completion through to the final ending.

\item \textbf{Design House Escape (5 milestones)}: Room escape game where a player solves creative design-themed puzzles to escape from a modern house. \underline{Milestone}: 
(1) Check whether a paper cube appears in the room with ``2'' marked on the wall,  
(2–4) Confirm whether three specific doors have been opened,
(5) Verify task completion through to the final ending.

\item \textbf{Paint Room Escape (8 milestones)}: Room escape game where a player uses art supplies and solves color-based puzzles to escape from a paint-themed room. \underline{Milestone}: 
(1–7) Confirm the color sequence of doors: red → light blue → orange → light blue (with pen present) → navy → yellow → green,  
(8) Verify task completion through to the final ending.

\item \textbf{Mirror Room Escape (5 milestones)}: Room escape game where a player solves reflection and light-based puzzles to escape from a room filled with mirrors. \underline{Milestone}:
(1) Check whether the colorful door has been opened,  
(2) Confirm opening of the glass bathroom door,  
(3) Verify that a message appears on the mirror when exposed to steam,  
(4) Check if the right-side door has been opened,  
(5) Verify task completion through to the final ending.

\item \textbf{Elevator Room Escape (4 milestones)}: Room escape game where a player must repair and operate an elevator to escape from a building. \underline{Milestone}:
(1–2) Confirm which floors are accessible via the elevator buttons (2nd and 3rd floors),  
(3) Verify the existence of a ceiling passage leading to the pulley room on the 3rd floor,  
(4) Verify task completion through to the final ending.

\item \textbf{Pico Sim Date (continuous score)}: Visual novel game where a player controls Pico, a boy who must win the affection of Nene, a girl he hopes to make his future wife, within 75 days. A Player balances relationship-building activities with battles against romantic rivals. \underline{Milestone}: 
Measure the affinity score shown next to the heart icon in the "Stats" menu.

\item \textbf{Festival Days Sim Date (continuous score)}: Visual novel game where a player has 30 days to get a boyfriend before the school festival, interacting with four unique boys and experiencing multiple endings. \underline{Milestone}:
Measure the maximum experience value among the four male characters, viewable via the "Stat" option in the Home menu.

\item \textbf{Kingdom Days (continuous score)}: Visual novel game set in a medieval kingdom, where a player, as Princess Rose, must build relationships with five characters in 30 days after fleeing her home and seeking refuge in a foreign land. \underline{Milestone}:
Measure the maximum experience value among the five male characters, accessible through the "Menu."

\item \textbf{Idol Days Sim Date (continuous score)}: Visual novel game in which a player has 30 days to date boys, improve skills, and prepare for a concert, aiming for love and success as an aspiring idol. \underline{Milestone}:
Measure the maximum experience value among the five male characters, accessible through the "Menu."

\item \textbf{Community College Sim (continuous score)}: Life simulation game where a player has 100 days to survive community college, balancing partying, studying, and socializing as a dropout, nerd, foreign kid, or grownup. \underline{Milestone}:
Measure the combined total of Intelligence and Social Skills, accessible via the "Stats" menu at the top of the screen.


\item \textbf{Sort the Court (continuous score)}: Management simulation game where a player acts as a king, making yes-or-no choices that affect the kingdom’s wealth, population, and happiness while managing events and relationships with citizens. \underline{Milestone}:
Measure the final values of population and happiness, then normalize each by the corresponding maximum values obtained via human walkthroughs; the final score is computed as the average of these normalized values.

\end{enumerate}
}

\subsection{Human GamePlay}
\label{subsec:appendix_human_game_play}
To evaluate human performance, we recruited 13 participants over the age of 18 who are fluent in English and have at least an undergraduate-level education.
Each participant reviewed and signed the consent form shown in Table~\ref{tab:consent_form}.
Before playing, the participants confirmed that they had never played the target games before.
They also completed a short questionnaire (Table~\ref{tab:qa_form}) assessing their prior experience or familiarity with classic adventure games.

\begin{table}[h!]
\scriptsize
\centering
\begin{tabular}{@{}p{\linewidth}@{}}
\toprule
\textbf{\benchmarkName Human Study Consent Form} \\
\midrule

This study involves collecting gameplay demonstrations from human participants for the \textit{\benchmarkName} benchmark. By signing this form, participants agree to the following:

~

1. \textbf{Study Overview}: Gameplay demonstrations, including input logs and screen recordings, will be used for game-playing AI research.

2. \textbf{Eligibility}: Participants must be 18 or older and provide informed, voluntary consent.

3. \textbf{Data Collection}: Collected data includes (a) mouse/keyboard input logs, (b) screen recordings with timestamps, and (c) questionnaire responses. Data will be used solely for research purposes.

4. \textbf{Anonymization \& Privacy}: All data are anonymized; no personally identifiable information will be disclosed.

5. \textbf{Voluntary Participation}: Participants may withdraw at any time and request data deletion without penalty.

6. \textbf{Risks}: No major risks are anticipated. Minor fatigue may occur due to screen exposure.

7. \textbf{Compensation}: Participants will receive \{national minimum wage\} per game.

8. \textbf{Play Duration}: Recommended play time per game is 15 to 90 minutes, regardless of game completion.

9. \textbf{Use of Results}: Results will be published in anonymized form for academic use. Participants may request access to final publications.

~

\textbf{Consent Statement:}  
I have read and understood the above. I voluntarily agree to participate and allow my data to be used for research purposes. \\

\bottomrule
\end{tabular}
\caption{Consent form provided to human participants.}
\label{tab:consent_form}
\end{table}

\begin{table}[h!]
\scriptsize
\centering
\begin{tabular}{@{}p{\linewidth}@{}}
\toprule
\textbf{Participant Background Questionnaire} \\
\midrule

\textbf{Q1. Familiarity with Classic Adventure Genres}  
\\
Have you previously played classic adventure games (\eg, Mystery/Detective, Hidden Object, Room Escape, Visual Novel, Life/Management Simulation)?  
For example, have you played escape room games or similar titles more than three times in the past three years?
If yes, please list any games you’ve played extensively or consider yourself skilled at.  
\\
~ \\

\textbf{Q2. General Gaming Experience}  
\\
Regardless of genre, are you generally familiar with playing video games?
If yes, please list any games you’ve played extensively or consider yourself skilled at.  
\\
~ \\

\textbf{Q3. Inexperience with Games}  
\\
Are you unfamiliar with video games or have little to no experience playing them?  
\\
\bottomrule
\end{tabular}
\caption{Pre-task questionnaire completed by each participant.}
\label{tab:qa_form}
\end{table}

The participants then used their own desktop/laptop to play the games.
They enabled input logging (\ie mouse and keyboard actions) and screen recording before starting each game.

After completing each game, the participants submitted their input logs and screen recordings to the authors.
We then evaluated the demonstrations based on predefined milestones.
For each game, we recorded the number of steps taken to reach each milestone.
In simulation-based games, we tracked the score at specific steps.

For each game, we collected gameplay demonstrations from three different participants.
On average, each participant completed approximately seven to eight games, with each game taking about 26 minutes.

\subsubsection{Statistics of Observation-Behavior Gap}
\label{subsubsec:appendix_obgap_stat}

To quantify the observation-behavior gap, we first focused on games from subgenres that feature discrete milestones. 
Specifically, we selected 7 games (2 mystery/detective, 2 hidden objects, and 3 room escape).
For each game, we randomly sampled 4 milestone-relevant clues that are essential for story progression.
Using gameplay logs from 3 human players, we measured the observation-behavior gap as the number of steps between when a clue was first observed and when it was acted upon.
We then computed the average gap across the players for each clue.

\begin{table*}[h!]
\centering
\begin{adjustbox}{width=\linewidth}
\begin{tabular}{lrrrrr}
\toprule
\textbf{Game} & \textbf{Gap \#1} & \textbf{Gap \#2} & \textbf{Gap \#3} & \textbf{Gap \#4} & \textbf{Average} \\
\midrule
Sherlock Holmes: The Tea Shop Murder Mystery & 95.3  & 561.7 & 321.7 & 655.3 & 408.5 $\pm$ 251.7 \\
Dakota Winchester's Adventures               & 113.7 & 322.0 & 211.0 & 229.3 & 219.0 $\pm$ 85.4  \\
Grim Tales: The Bride                        & 57.0  & 62.0  & 72.0  & 91.7  & 70.7 $\pm$ 15.3   \\
Grim Tales: The Legacy Collector's Edition   & 58.3  & 78.7  & 95.7  & 122.0 & 88.7 $\pm$ 27.0   \\
Computer Office Escape                       & 1001.3 & 584.3 & 120.3 & 47.3  & 438.3 $\pm$ 444.3 \\
Camping Room Escape                          & 307.0 & 632.3 & 150.3 & 57.3  & 286.8 $\pm$ 252.4 \\
Space Museum Escape                          & 303.0 & 157.3 & 240.0 & 284.0 & 246.1 $\pm$ 64.8  \\
\bottomrule
\end{tabular}
\end{adjustbox}
\caption{Observation-behavior gap analysis across games. Values show four gaps and their average $\pm$ standard deviation.}
\label{tab:obgap_analysis}
\end{table*}

The overall average observation-behavior gap in Table~\ref{tab:obgap_analysis} is \textbf{251.1$\pm$142.1} steps, which supports the ``substantial'' step gaps discussed in \S~\ref{subsec:human_game_play}.
This value is nearly five times greater than the average human playtime in VisEscape~\citep{lim2025visescape} (52.8 steps).
We also found genre-specific trends: (1) Mystery/detective and room escape games showed larger and more variable gaps across games. (2) Hidden object games had shorter average playtimes and correspondingly smaller, more consistent gaps between games.

\begin{figure}[h!]
    \centering
    \includegraphics[width=\columnwidth]{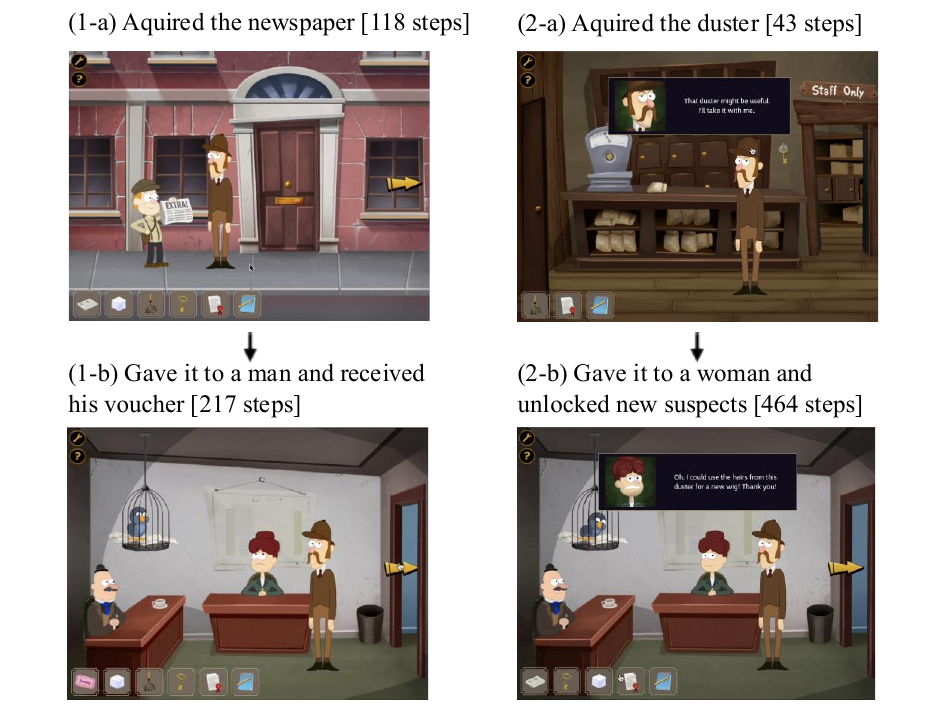}
    \caption{Examples illustrating the long-term observation-behavior gap from human player demonstrations in \textit{Sherlock Holmes: The Tea Shop Murder Mystery}. (1-a) The player acquired a newspaper after 118 steps. (1-b) The player gave the newspaper to a man and received a voucher after 217 steps, resulting in a step gap of 99 between acquiring the newspaper and receiving the voucher. (2-a) The player acquired a duster after 43 steps. (2-b) The player gave the duster to a woman, unlocking new suspects after 464 steps, with a step gap of 421 between acquiring the duster and unlocking new suspects.}
    \label{fig:obgap_qual}
\end{figure}

Additionally, we provide examples from the game \textit{Sherlock Holmes: The Tea Shop Murder Mystery} illustrating the long-term observation-behavior gap from human player data in Figure~\ref{fig:obgap_qual}.

\subsection{CUA-as-a-Judge}
\label{subsec:appendix_cua_as_a_judge}

\begin{figure}[h!]
    \centering
    \includegraphics[width=\columnwidth]{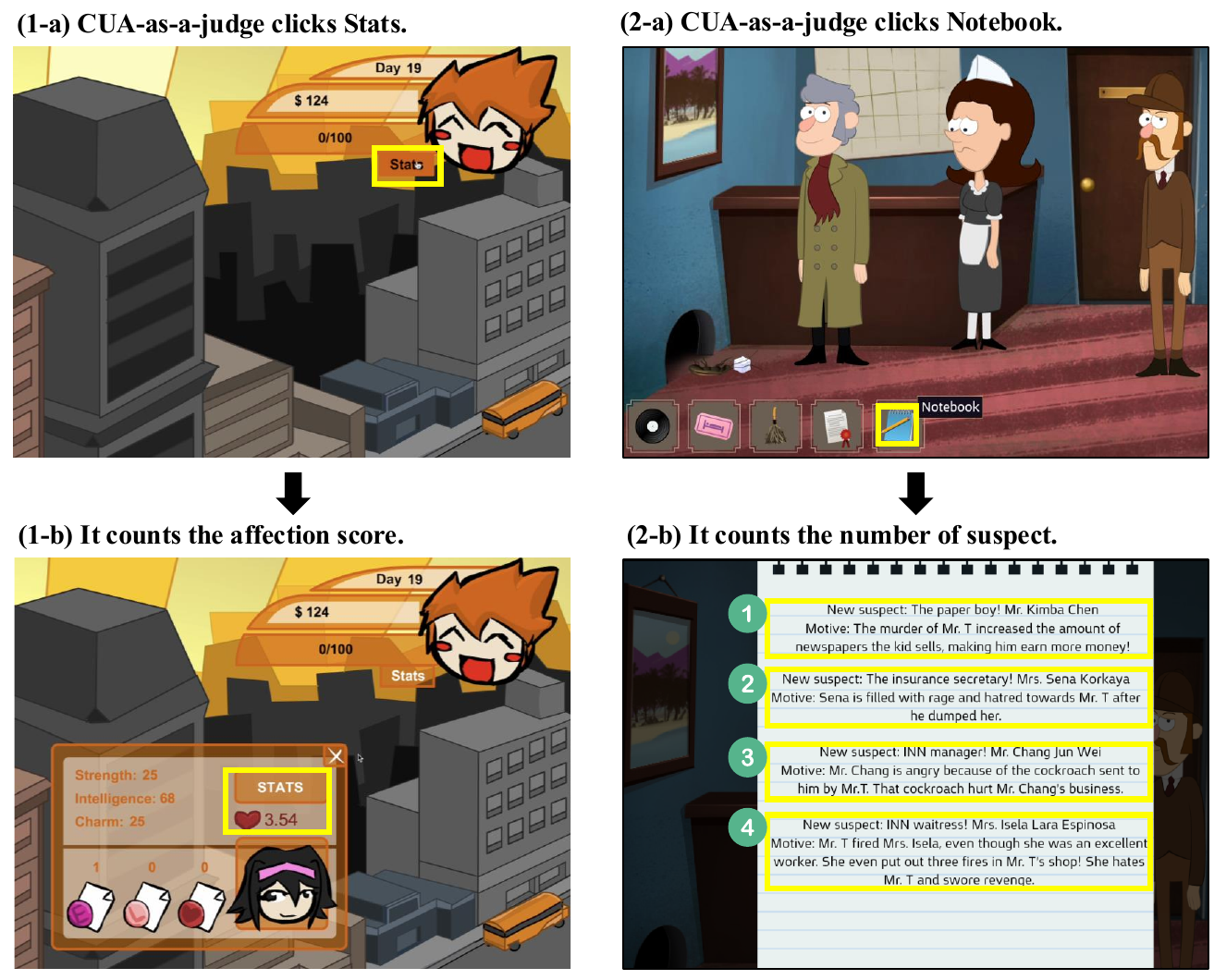}
    \caption{CUA-as-a-Judge verifies game progress by interacting with the environment. Left: (1-a) The judge clicks the ``Stats'' button in \textit{Pico Sim Date} to (1-b) verify the character's affection score. Right: (2-a) The judge clicks the ``Notebook'' item in \textit{Sherlock Holmes: The Tea Shop Murder Mystery} to (2-b) count the number of suspects (5) identified by the player, confirming milestone completion.}
    \label{fig:cua_as_a_judge_example}
\end{figure}

We provide Figure~\ref{fig:cua_as_a_judge_example} as an example of CUA-as-a-Judge automatically evaluating game progression by interacting with game interfaces to verify milestone completion through checking affection scores and suspect counts.

\noindent
\textbf{Milestone structure and evaluation strategy.}
Games in \benchmarkName differ in how milestones can be observed and assessed.
Some follow a step-by-step structure, where completion can only be confirmed by sequentially checking each individual event or condition (\eg whether a specific interaction occurred or an NPC disappeared). In these 11 games, CUA-as-a-Judge verifies milestones one by one, halting at the first unmet milestone and reporting progress as $K_\text{m} / N_\text{m}$, as described in \S~\ref{subsec:cua_as_a_judge}.

In contrast, 23 games, including visual novel and simulation games, expose cumulative game state (\eg number of suspects unlocked, locations discovered, affection scores), allowing all $N_\text{m}$ milestones to be assessed in a single pass. In these cases, the final score reflects the total number of milestones completed.

\noindent  
\textbf{Evaluation agreement vs. human judge.}  
For each of the \numGames games, we evaluate CUA-as-a-Judge on 8 or 9 sampled evaluation instances.
If a game has 4 milestones, for example, the possible milestone completions range from 0 to 4 (\ie 5 levels).
To account for variations in final gameplay states (even when the milestone count is the same), we construct two distinct examples per level, yielding up to 10 evaluation instances per game.
From these, we randomly sample 8 or 9 cases and assess whether CUA-as-a-Judge correctly estimates the milestone completion level, resulting in a total of 300 evaluation samples across all games.

\begin{table}[h!]
\scriptsize
\centering
\begin{adjustbox}{width=\columnwidth}
\begin{tabular}{lc}
\toprule
\textbf{Game} & \textbf{Accuracy} \\
\midrule
Sherlock Holmes: The Tea Shop Murder Mystery & 9/9 \\
Sherlock Holmes 2 & 6/8\\
Vortex Point 1 & 9/9\\
Vortex Point 2 & 8/9\\
Vortex Point 3 & 9/9\\
Pierre Hotel & 9/9\\
Small Town Detective & 9/9\\
Dakota Winchester’s Adventures & 8/9\\
Saucy Devil Gordon & 8/9\\
Ray and Cooper 2 & 9/9\\
Nick Bounty: A Case of the Crabs & 8/8\\
Grim Tales: The Bride & 7/9\\
Grim Tales: The Legacy Collector’s Edition & 7/9\\
Computer Office Escape & 9/9\\
Crimson Room & 6/8\\
Camping Room Escape & 9/9\\
Chemical Room Escape & 9/9\\
Space Museum Escape & 9/9\\
Vending Machine Room & 6/8\\
Wood Workshop Escape & 6/8\\
Geometric Room Escape & 9/9\\
Game Cafe Escape & 9/9\\
Machine Room Escape & 9/9\\
VideoStudio Escape & 9/9\\
Design House Escape & 9/9\\
Paint Room Escape & 9/9\\
Mirror Room Escape & 8/9\\
Elevator Room Escape & 6/8\\
Pico Sim Date & 9/9\\
Festival Days Sim Date & 9/9\\
Kingdom Days & 9/9\\
Idol Days Sim Date & 9/9\\
Community College Sim & 9/9\\
Sort the Court & 9/9\\
\midrule
\textbf{Total} & 282/300\\
\bottomrule
\end{tabular}
\end{adjustbox}
\caption{Per-game accuracy of CUA-as-a-Judge compared to human judgment.}
\label{tab:per_game_accuracy}
\end{table}

Our comparison shows a high agreement, with an accuracy of 94.00\%, Spearman correlation of 0.9912, and Pearson correlation of 0.9999.
Per-game accuracy results are presented in Table~\ref{tab:per_game_accuracy}.

\section{Details on Experiments}
\subsection{Implementation Details}
\label{subsec:appendix_implementation_details}
GPT-4o (\texttt{gpt-4o-2024-08-06}), Claude-3.7-Sonnet (\texttt{claude-3-7-sonnet-20250219}), OpenAI CUA (\texttt{computer-use-preview-2025-03-11}), and Claude-3.7-Sonnet Computer-Use (\texttt{claude-3-7-sonnet-20250219} with tool version \texttt{20250124}) were all accessed via proprietary APIs using default hyperparameters.
All experiments were conducted on 28 personal computers capable of running the Flash games (7 Macs and 21 Windows). 
For UGround-V1-7B and UI-TARS-1.5-7B, we ran them on a single NVIDIA RTX A6000 GPU and accessed the APIs via VLLM~\citep{kwon2023vllm}.
Each experiment typically took about 12 to 18 hours per game.

For all GUI agents, the most recent 10 trajectories were explicitly provided to the agent. Beyond this, we made efforts to preserve each baseline's original configuration as much as possible. Some baselines~\citep{tan2024cradle,agashe2025agents2,qin2025uitars} implemented retrieval or summary modules to realize long-term memory, while for OpenAI CUA and Claude-3.7-Sonnet Computer-Use, the internal mechanisms are not publicly known, so whether they implement long-term memory remains unclear.

For our \methodName framework, we use Claude-3.7-Sonnet Computer-Use for Clue Seeker and Problem Solver, and Claude-3.7-Sonnet for Clue Mapper.
For the hyperparameters \{$N_\text{seek}$,$N_\text{solve}$\}, we set them to (5,2) for visual novels and simulation genres, and (15,5) for all others.
We set $K=5$ for Clue Mapper.

Due to the high API cost, all experiments were conducted as single runs.

\subsubsection{Detailed Prompts}
\label{subsubsec:appendix_detailed_prompts}
We provide detailed prompts of (1) CUA-as-a-Judge, (2) baseline agents, and (3) \methodName for the game ‘Sherlock Holmes: The Tea Shop Murder Mystery’ as an example:

\begin{tcolorbox}[title=CUA-as-a-Judge Prompt, breakable]
\small
Your task is to locate and click the blue notebook icon in the game interface (usually found at the far right of the inventory bar). 
Once the notebook is opened, do not perform any additional actions. 
Carefully read the displayed text (do not scroll) and count how many times the phrase \texttt{New Suspect} appears in the note. 
If there is nothing written in the note, it means there is no new suspect.
Once counted, output the result in the following format:

~

\textbf{Output Format}

\texttt{New Suspect: [Num of occurrences]}

Do not perform any further interactions after counting. 
Your task ends once the count is provided.
\end{tcolorbox}

\begin{tcolorbox}[title=Basic Prompt (for baseline agents), colback=gray!10, colframe=red!70!black, breakable]
\small
\textbf{[Instruction]} \\
You’re a game agent solving an adventure game. Adventure games can involve variables and often require lateral thinking and creative problem-solving. Rather than focusing solely on the ultimate goal, try to solve problems using common sense and imaginative thinking. \\[0.5em]

\textbf{[Important Notice]} \\
To ensure a smooth experience and prevent unexpected issues, please read and follow the instructions below carefully:
\begin{itemize}
    \item Do not repeatedly interact with objects that may not be clickable. Even if nothing happens, your action might have already been registered-or the object may simply not be interactive.
    \item Avoid repeating the same action excessively. A lack of visible response or an unexpected result does not mean the action failed.
    \item Do not click the setup or question buttons in the top-left corner of the screen.
    \item Do not leave the game screen.
    \item Do not access Settings or Help:
    \begin{itemize}
        \item Do not click the wrench icon (Settings).
        \item Do not click the question mark icon (Help or Hints).
    \end{itemize}
\end{itemize}

\textbf{[Prompt]} \\
You are now the lead agent in ``Sherlock Holmes: The Tea Shop Murder Mystery’’ - a point-and-click detective game. Everything in this game is controlled with simple mouse clicks. Search the tea shop carefully, interact with objects, uncover hidden clues, and solve the murder mystery. \\[0.5em]

\textbf{[Basic Rules]}
\begin{itemize}
    \item Observe carefully. Don’t rush or search randomly - every detail could matter. What seems ordinary might hide secrets.
    \item Think logically and laterally. Don’t brute-force or guess wildly. Follow the trail of clues using reason, insight, and creativity.
    \item Investigate step by step. Whether solving a mystery or unlocking a puzzle, progress comes from careful observation and deduction.
    \item Trust your instincts. Sometimes the answer is hidden in plain sight - don’t overlook the obvious.
\end{itemize}

\textbf{[Information]} \\
Mr. T has been murdered by someone. But we don't know who the criminal is. \\
Mr. T was the owner of the local tea shop. \\
\ldots \\[0.5em]

\textbf{[Completion Condition]} \\
The game is considered complete only when you have gathered enough evidence, logically identified the culprit, and correctly selected the true suspect during the final decision. A clear message or scene will confirm that the case has been solved. \\[0.5em]

\textbf{[Completion Signal]} \\
Once you’ve solved the case, type the following to signal completion: \\
\texttt{[Done]}

\end{tcolorbox}

\begin{tcolorbox}[title=COAST: Clue Seeker Prompt, 
colback=gray!10, colframe=blue!70!black, breakable]
\small
\textbf{\{Basic Prompt\}}

\vspace{0.5em}
\textbf{[Action Prompt]} \\
Your job is to extract all clues visible on the screen and summarize what you observe.

\vspace{0.5em}
\textbf{[Expected Behavior]} \\
Return the following:
\begin{itemize}
    \item A list of clues found. Each clue should include:
    \begin{itemize}
        \item \texttt{clue}: short name or label for the clue
        \item \texttt{description}: what the clue seems to represent or imply
        \item \texttt{location}: where the clue was found - include both the specific spot and a short description of the surrounding environment
        \item \texttt{type}: categorize the clue using one of the following:
        \begin{itemize}
            \item \texttt{item}: Tools or objects the player can collect or interact with
            \item \texttt{note}: Written information such as signs, notes, or documents
            \item \texttt{code}: Visible numbers, passcodes, symbols, or puzzle sequences
            \item \texttt{visual cue}: Visual cues like arrows, lighting, gaze direction, or environmental emphasis
            \item \texttt{status}: UI state indicators
            \item \texttt{conversation}: Any meaningful dialogue or internal monologue text shown on screen
        \end{itemize}
        \item \texttt{interactable}: true if the player can interact with this clue, false otherwise
        \item \texttt{usage\_hint}: how this clue might be used or why it is important
    \end{itemize}
    \item A short summary of the current observation-action pair (episodic memory).
    Each memory should include:
    \begin{itemize}
        \item \texttt{action}: what you did and what was observed as a result
        \item \texttt{place}: what the current area/room looks like, its notable features
    \end{itemize}
\end{itemize}

\vspace{0.5em}
\textbf{[Result Format]} \\
Respond in this exact JSON format, wrapped in \texttt{<RESPO>} tags, like this:

\begin{verbatim}
<RESPO>  
{  
  "clues": [  
    {  
      "clue": "<Short name or label for the clue>",  
      "description": "<What the clue seems to represent or imply>",  
      "location": "<Specific spot + surrounding context>",  
      "type": "<item | note | code | visual cue | status | conversation>",  
      "interactable": <true | false>,  
      "usage_hint": "<How this clue might be used or why it could be important>"  
    }  
  ],  
  "episodic_memory": [  
    {  
      "action": "<What the player did and what was observed as a result>",  
      "place": "<Description of the room or environment where the action occurred>"  
    }  
  ]  
}  
</RESPO>
\end{verbatim}

If your response does not strictly follow this format, it will be discarded.  
Do not store the same clue more than once in memory.

\vspace{0.5em}
\textbf{[Clues]} \\
\{clues\}

\end{tcolorbox}

\begin{tcolorbox}[title=COAST: Clue Mapper Prompt,
  colback=gray!10, colframe=blue!70!black,
  breakable]
\small
\textbf{\{Basic Prompt\}}

\vspace{0.5em}
\textbf{[Action Prompt]} \\
You are a reasoning agent matching current clues with episodic memory from past gameplay.
Your goal is to find meaningful --- and possibly non-obvious --- connections between clues and past events using:
\begin{itemize}
  \item \textit{abductive reasoning}: inferring the most plausible explanation from incomplete or ambiguous information
  \item \textit{lateral thinking}: creative, indirect associations beyond surface similarity
\end{itemize}

\noindent For each clue:
\begin{itemize}
  \item Identify an episodic memory where the clue could plausibly have helped --- even if the connection is indirect or interpretive.
  \item Determine the concrete action the player should now take based on that match.
\end{itemize}

\vspace{0.5em}
\textbf{[Expected behavior]} 
\begin{itemize}
  \item Use each clue's \texttt{description}, \texttt{type}, and \texttt{usage\_hint} to inform your reasoning.
  \item Go beyond surface-level similarity --- prioritize plausible, creative mappings.
  \item Favor abductive reasoning --- what might this clue explain or reveal?
  \item Explore lateral connections --- metaphorical, thematic, or functional.
  \item Only match when a memory clearly presents a situation where the clue could have been helpful.
  \item Be specific and grounded. If uncertain, omit the match.
  \item Return up to 5 of the most insightfully plausible matches --- quality over quantity.
  \item If no valid matches are found, return:
\end{itemize}

\begin{verbatim}
<RESPO>[Nobody]</RESPO>
\end{verbatim}

\noindent Do not fabricate connections.

\vspace{0.5em}
\textbf{[Result Format]} \\
Respond in JSON format like this:
\begin{verbatim}
<RESPO>  
[  
  {  
    "clue": {  
      "name": "<short clue name>",  
      "description": "<what the clue seems to represent or imply>",  
      "location": "<specific spot + environment context>",  
      "type": "<item | note | code | visual cue | status | conversation>",  
      "interactable": <true | false>,  
      "usage_hint": "<how this clue might be used or why it could be important>"  
    },  
    "related_memory": "<a specific past observation where this clue would have been useful>",  
    "expected_action": "<concrete action the player should now take using this clue>"  
  }  
]  
</RESPO>
\end{verbatim}
If the format is not strictly followed, the response will be discarded.

\vspace{0.5em}
\textbf{[Clues]} \\
\{clues\}

\vspace{0.5em}
\textbf{[Episodic Memory]} \\
\{episodic memory\}

\vspace{0.5em}
Do not generate mapping memory that has already succeeded.

\vspace{0.5em}
\textbf{[Success Memory]} \\
\{success memory\}

\end{tcolorbox}

\begin{tcolorbox}[title=COAST: Problem Solver Prompt,
  colback=gray!10, colframe=blue!70!black,
  breakable]
\small
\textbf{\{Basic Prompt\}}

\vspace{0.5em}
\textbf{[Action Prompt]} \\
You are now trying to solve games using previously discovered clues and their related observations.
Each clue is paired with a related episodic memory from your past exploration.
Use this information to decide on the most logical and effective next action to progress in the game.

\noindent Each episodic memory is structured as:
\begin{itemize}
  \item \texttt{action}: what you did and what you observed
  \item \texttt{place}: description of the environment or scene
\end{itemize}

\vspace{0.5em}
\textbf{[Expected behavior]}
\begin{itemize}
  \item Choose a clear goal based on the clue-to-memory mapping you are given.
  \item Take a meaningful action in the game to pursue that goal.
  \item Summarize what happened after the action.
  \item Return the result in the format below:
  \begin{itemize}
    \item \texttt{episodic\_memory}: what happened during this step (at least one item)
    \item \texttt{mapping\_result} (optional), if your action clearly relates to a clue, include:
    \begin{itemize}
      \item \texttt{goal}: what problem you were trying to solve
      \item \texttt{reasoning}: why that clue and memory were relevant to the goal
      \item \texttt{result}: one of \texttt{"Success"} or \texttt{"Fail"} depending on whether your action clearly used the clue to solve a problem
    \end{itemize}
  \end{itemize}
\end{itemize}

\vspace{0.5em}
\textbf{[Result Format]} \\
Respond in JSON format like this:
\begin{verbatim}
<RESPO>  
{  
  "episodic_memory": [  
    {  
      "action": "<What the player did and what was observed as a result>",  
      "place": "<Description of the room or area where it happened>"  
    }  
  ],  
  "mapping_result": [  
    {  
      "clue": "<Clue name used in this action>",  
      "related_memory": "<The relevant episodic memory entry this clue connects to>",  
      "goal": "<What the player was trying to achieve by using the clue>",  
      "reasoning": "<Why this clue and memory logically support that goal>"  
    }  
  ],  
  "result": "<Success | Fail>"  
}  
</RESPO>
\end{verbatim}
If your response does not strictly follow the format, it will be discarded.

\vspace{0.5em}
\textbf{[Important about Success]} \\
An action is only considered a \texttt{"Success"} if the clue was effectively used to solve a specific puzzle or problem --- for example, using a pattern from books on a shelf to open a secret compartment based on a past observation.
Simply interacting with objects is not enough.

\noindent To qualify as a true \texttt{Success}, the outcome must include a meaningful in-game change, such as:
\begin{itemize}
  \item Obtaining an item
  \item Unlocking a new area
  \item Updating a stat
  \item Triggering story progression
\end{itemize}

\vspace{0.5em}
\textbf{[Action Process]}
\begin{itemize}
  \item Select a goal based on the mapping between clue and memory. Explain why this goal is relevant.
  \item Act accordingly in the game world.
  \item At the final turn, assess the outcome.
  \item If the clue helped solve a problem, it’s a \texttt{"Success"}; otherwise, it’s a \texttt{"Fail"}.
  \item When selecting a goal, be sure to reference the clue’s metadata:
  \begin{itemize}
    \item \texttt{type}: the kind of clue (\eg, item, code, note, etc.)
    \item \texttt{interactable}: whether the player can use it
    \item \texttt{usage\_hint}: what the clue suggests it might be useful for
  \end{itemize}
\end{itemize}

\vspace{0.5em}
\textbf{[Mapping History]} \\
Clue: \{clue 1\} \\
Related Memory: \{related memory 1\} \\
Expected Action: \{expected action 1\} \\
\ldots

\end{tcolorbox}

\begin{table*}[h!]
\centering
\resizebox{\linewidth}{!}{%
\begin{tabular}{lccccccccccccccc}
\toprule
\multirow{2}{*}{Game} & 
\multicolumn{3}{c}{\makecell{OpenAI\\CUA}} & 
\multicolumn{3}{c}{\makecell{Claude-3.7\\Computer-Use}} & 
\multicolumn{3}{c}{\makecell{Claude-3.7\\Computer-Use\\+\methodName(Ours)}} & 
\multicolumn{3}{c}{\makecell{Human\\(max 1K steps)}} & 
\multicolumn{3}{c}{\makecell{Human\\(unlimited)}} \\
\cmidrule(lr){2-4} \cmidrule(lr){5-7} \cmidrule(lr){8-10} \cmidrule(lr){11-13} \cmidrule(lr){14-16}
& \makecell{SR\\(\%)} & \makecell{MCR\\(\%)} & \makecell{Stp\\(\#)} 
& \makecell{SR\\(\%)} & \makecell{MCR\\(\%)} & \makecell{Stp\\(\#)}
& \makecell{SR\\(\%)} & \makecell{MCR\\(\%)} & \makecell{Stp\\(\#)}
& \makecell{SR\\(\%)} & \makecell{MCR\\(\%)} & \makecell{Stp\\(\#)}
& \makecell{SR\\(\%)} & \makecell{MCR\\(\%)} & \makecell{Stp\\(\#)} \\
\midrule

\multicolumn{16}{l}{\textbf{Point-and-Click Adventure (Mystery/Detective)}} \\
Sherlock Holmes: The Tea Shop Murder Mystery & 0.0 & 50.0 & 1000 & 0.0 & 16.7 & 1000 & 0.0 & 50.0 & 1000 & 100.0 & 100.0 & 636.6 & 100.0 & 100.0 & 636.6 \\
Sherlock Holmes 2 & 0.0 & 37.5 & 1000 & 0.0 & 37.5 & 1000 & 0.0 & 62.5 & 1000 & 66.7 & 95.8 & 740.7 & 100.0 & 100.0 & 753.0 \\
Vortex Point 1 & 0.0 & 0.0 & 1000 & 0.0 & 0.0 & 1000 & 0.0 & 0.0 & 1000 & 66.7 & 81.0 & 833.7 & 100.0 & 100.0 & 987.3 \\
Vortex Point 2 & 0.0 & 0.0 & 1000 & 0.0 & 0.0 & 1000 & 0.0 & 0.0 & 1000 & 66.7 & 86.7 & 747.0 & 100.0 & 100.0 & 796.0 \\
Vortex Point 3 & 0.0 & 20.0 & 1000 & 0.0 & 0.0 & 1000 & 0.0 & 20.0 & 1000 & 100.0 & 100.0 & 519.0 & 100.0 & 100.0 & 519.0 \\
Pierre Hotel & 0.0 & 0.0 & 1000 & 0.0 & 0.0 & 1000 & 0.0 & 0.0 & 1000 & 66.7 & 88.9 & 726.7 & 100.0 & 100.0 & 731.0 \\
Small Town Detective & 0.0 & 0.0 & 1000 & 0.0 & 0.0 & 1000 & 0.0 & 0.0 & 1000 & 66.7 & 94.4 & 780.7 & 100.0 & 100.0 & 815.0 \\
Dakota Winchester’s Adventures & 0.0 & 0.0 & 1000 & 0.0 & 0.0 & 1000 & 0.0 & 0.0 & 1000 & 100.0 & 100.0 & 562.7 & 100.0 & 100.0 & 562.7 \\
Saucy Devil Gordon & 0.0 & 0.0 & 1000 & 0.0 & 0.0 & 1000 & 0.0 & 0.0 & 1000 & 100.0 & 100.0 & 379.7 & 100.0 & 100.0 & 379.7 \\
Ray and Cooper 2 & 0.0 & 0.0 & 1000 & 0.0 & 0.0 & 1000 & 0.0 & 0.0 & 1000 & 100.0 & 100.0 & 952.5 & 100.0 & 100.0 & 952.5 \\
Nick Bounty: A Case of the Crabs & 0.0 & 20.0 & 1000 & 0.0 & 20.0 & 1000 & 0.0 & 20.0 & 1000 & 33.3 & 80.0 & 843.7 & 100.0 & 100.0 & 1161.0 \\
\midrule

\multicolumn{16}{l}{\textbf{Hidden Object}} \\
Grim Tales: The Bride & 100.0 & 100.0 & 274 & 0.0 & 91.7 & 1000 & 100.0 & 100.0 & 225 & 100.0 & 100.0 & 91.7 & 100.0 & 100.0 & 91.7 \\
Grim Tales: The Legacy Collector’s Edition & 100.0 & 100.0 & 164 & 0.0 & 75.0 & 1000 & 100.0 & 100.0 & 647 & 100.0 & 100.0 & 121.0 & 100.0 & 100.0 & 121.0 \\
\midrule

\multicolumn{16}{l}{\textbf{Room Escape}} \\
Computer Office Escape & 0.0 & 0.0 & 1000 & 0.0 & 0.0 & 1000 & 0.0 & 0.0 & 1000 & 66.7 & 80.0 & 851.3 & 100.0 & 100.0 & 987.0 \\
Crimson Room & 0.0 & 35.7 & 1000 & 0.0 & 35.7 & 1000 & 0.0 & 28.6 & 1000 & 0.0 & 64.3 & 1000.0 & 100.0 & 100.0 & 3133.0 \\
Camping Room Escape & 0.0 & 22.2 & 1000 & 0.0 & 22.2 & 1000 & 0.0 & 44.4 & 1000 & 0.0 & 44.4 & 1000.0 & 100.0 & 100.0 & 1734.0 \\
Chemical Room Escape & 0.0 & 25.0 & 1000 & 0.0 & 25.0 & 1000 & 0.0 & 25.0 & 1000 & 0.0 & 79.2 & 1000.0 & 100.0 & 100.0 & 1303.3 \\
Space Museum Escape & 0.0 & 0.0 & 1000 & 0.0 & 0.0 & 1000 & 0.0 & 0.0 & 1000 & 33.3 & 77.8 & 994.3 & 100.0 & 100.0 & 1128.3 \\
Vending Machine Room & 0.0 & 22.2 & 1000 & 0.0 & 33.3 & 1000 & 0.0 & 44.4 & 1000 & 100.0 & 100.0 & 664.3 & 100.0 & 100.0 & 664.3 \\
Wood Workshop Escape & 0.0 & 14.3 & 1000 & 0.0 & 42.9 & 1000 & 0.0 & 42.9 & 1000 & 33.3 & 90.5 & 949.3 & 100.0 & 100.0 & 1104.0 \\
Geometric Room Escape & 0.0 & 0.0 & 1000 & 0.0 & 0.0 & 1000 & 0.0 & 0.0 & 1000 & 33.3 & 55.5 & 882.3 & 100.0 & 100.0 & 1371.7 \\
Game Cafe Escape & 0.0 & 0.0 & 1000 & 0.0 & 0.0 & 1000 & 0.0 & 0.0 & 1000 & 33.3 & 73.3 & 933.3 & 100.0 & 100.0 & 993.7 \\
Machine Room Escape & 0.0 & 0.0 & 1000 & 0.0 & 0.0 & 1000 & 0.0 & 0.0 & 1000 & 100.0 & 100.0 & 819.3 & 100.0 & 100.0 & 819.3 \\
VideoStudio Escape & 0.0 & 0.0 & 1000 & 0.0 & 0.0 & 1000 & 0.0 & 0.0 & 1000 & 0.0 & 25.0 & 1000.0 & 100.0 & 100.0 & 1697.7 \\
Design House Escape & 0.0 & 0.0 & 1000 & 0.0 & 0.0 & 1000 & 0.0 & 0.0 & 1000 & 33.3 & 86.7 & 971.1 & 100.0 & 100.0 & 1183.7 \\
Paint Room Escape & 0.0 & 0.0 & 1000 & 0.0 & 0.0 & 1000 & 0.0 & 0.0 & 1000 & 0.0 & 91.7 & 1000.0 & 100.0 & 100.0 & 1171.7 \\
Mirror Room Escape & 0.0 & 0.0 & 1000 & 0.0 & 0.0 & 1000 & 0.0 & 0.0 & 1000 & 0.0 & 53.3 & 1000.0 & 100.0 & 100.0 & 2352.7 \\
Elevator Room Escape & 0.0 & 0.0 & 1000 & 0.0 & 0.0 & 1000 & 0.0 & 0.0 & 1000 & 0.0 & 66.7 & 1000.0 & 100.0 & 100.0 & 1161.7 \\
\midrule

\multicolumn{16}{l}{\textbf{Visual Novel (Dating Sim)}} \\
Pico Sim Date & 0.0 & 0.1 & 1000 & 0.0 & 4.6 & 1000 & 0.0 & 0.9 & 1000 & 0.0 & 13.0 & 1000.0 & 33.3 & 100.0 & 1751.0 \\
Festival Days Sim Date & 0.0 & 15.7 & 1000 & 0.0 & 0.6 & 1000 & 0.0 & 11.2 & 1000 & 0.0 & 41.7 & 1000.0 & 100.0 & 100.0 & 2798.7 \\
Kingdom Days & 0.0 & 2.2 & 1000 & 0.0 & 19.5 & 1000 & 0.0 & 1.5 & 1000 & 100.0 & 100.0 & 958.0 & 100.0 & 100.0 & 958.0 \\
Idol Days Sim Date & 0.0 & 17.4 & 1000 & 0.0 & 69.8 & 760 & 0.0 & 25.6 & 1000 & 100.0 & 100.0 & 873.0 & 100.0 & 100.0 & 873.0 \\
\midrule

\multicolumn{16}{l}{\textbf{Simulation}} \\
 Sort the Court & 0.0 & 37.4 & 1000 & 0.0 & 78.1 & 1000 & 0.0 & 95.5 & 1000 & 33.3 & 88.1 & 894.0 & 66.7 & 100.0 & 1145.7 \\
Community College Sim & 0.0 & 3.2 & 1000 & 0.0 & 9.2 & 1000 & 0.0 & 3.9 & 1000 & 0.0 & 27.4 & 1000.0 & 100.0 & 100.0 & 1992.0 \\
\midrule
\textbf{Average} & \textbf{5.9} & 15.4 & 954.1 & 0.0 & \underline{17.1} & 992.9 & \textbf{5.9} & \textbf{19.9} & 966.8 & 51.0 & 79.0 & 815.5 & 97.1 & 100.0 & 1142.0 \\
\bottomrule
\end{tabular}
}
\caption{Detailed results for all \numGames games across three GUI agents (\ie (1) OpenAI CUA, (2) Claude-3.7-Sonnet Computer-Use, and (3) Claude-3.7-Sonnet Computer-Use with \methodName) and human baselines.}
\label{tab:detailed_results_34}
\end{table*}

\begin{table*}[h!]
\centering
\resizebox{\linewidth}{!}{%
\begin{tabular}{lccccccccccccccc}
\toprule
\multirow{2}{*}{Game} & 
\multicolumn{3}{c}{\makecell{Claude-3.7 +\\UGround + Cradle}} & 
\multicolumn{3}{c}{\makecell{Claude-3.7 +\\Claude-3.7 + Cradle}} & 
\multicolumn{3}{c}{\makecell{Claude-3.7 +\\Claude-3.7 + Agent S2}} & 
\multicolumn{3}{c}{\makecell{GPT-4o +\\UGround + Cradle}} & 
\multicolumn{3}{c}{\makecell{UI-TARS-1.5-7B}} \\
\cmidrule(lr){2-4} \cmidrule(lr){5-7} \cmidrule(lr){8-10} \cmidrule(lr){11-13} \cmidrule(lr){14-16}
& \makecell{SR\\(\%)} & \makecell{MCR\\(\%)} & \makecell{Stp\\(\#)} 
& \makecell{SR\\(\%)} & \makecell{MCR\\(\%)} & \makecell{Stp\\(\#)}
& \makecell{SR\\(\%)} & \makecell{MCR\\(\%)} & \makecell{Stp\\(\#)}
& \makecell{SR\\(\%)} & \makecell{MCR\\(\%)} & \makecell{Stp\\(\#)}
& \makecell{SR\\(\%)} & \makecell{MCR\\(\%)} & \makecell{Stp\\(\#)} \\
\midrule

\multicolumn{16}{l}{\textbf{Point-and-Click Adventure (Mystery/Detective)}} \\
Sherlock Holmes: The Tea Shop Murder Mystery & 0.0 & 0.0 & 1000 & 0.0 & 0.0 & 1000 & 0.0 & 0.0 & 1000 & 0.0 & 0.0 & 1000 & 0.0 & 0.0 & 1000 \\
Sherlock Holmes 2 & 0.0 & 0.0 & 1000 & 0.0 & 12.5 & 1000 & 0.0 & 0.0 & 1000 & 0.0 & 0.0 & 1000 & 0.0 & 12.5 & 1000 \\
Vortex Point 1 & 0.0 & 0.0 & 1000 & 0.0 & 0.0 & 1000 & 0.0 & 0.0 & 1000 & 0.0 & 0.0 & 1000 & 0.0 & 0.0 & 1000 \\
Vortex Point 2 & 0.0 & 0.0 & 1000 & 0.0 & 0.0 & 1000 & 0.0 & 0.0 & 1000 & 0.0 & 0.0 & 1000 & 0.0 & 0.0 & 1000 \\
Vortex Point 3 & 0.0 & 20.0 & 1000 & 0.0 & 0.0 & 1000 & 0.0 & 0.0 & 1000 & 0.0 & 20.0 & 1000 & 0.0 & 0.0 & 1000 \\
Pierre Hotel & 0.0 & 0.0 & 1000 & 0.0 & 0.0 & 1000 & 0.0 & 0.0 & 1000 & 0.0 & 0.0 & 1000 & 0.0 & 0.0 & 1000 \\
Small Town Detective & 0.0 & 0.0 & 1000 & 0.0 & 0.0 & 1000 & 0.0 & 0.0 & 1000 & 0.0 & 0.0 & 1000 & 0.0 & 0.0 & 1000 \\
Dakota Winchester’s Adventures & 0.0 & 0.0 & 1000 & 0.0 & 0.0 & 1000 & 0.0 & 0.0 & 1000 & 0.0 & 0.0 & 1000 & 0.0 & 0.0 & 1000 \\
Saucy Devil Gordon & 0.0 & 0.0 & 1000 & 0.0 & 0.0 & 1000 & 0.0 & 0.0 & 1000 & 0.0 & 0.0 & 1000 & 0.0 & 0.0 & 1000 \\
Ray and Cooper 2 & 0.0 & 0.0 & 1000 & 0.0 & 0.0 & 1000 & 0.0 & 0.0 & 1000 & 0.0 & 0.0 & 1000 & 0.0 & 0.0 & 1000 \\
Nick Bounty: A Case of the Crabs & 0.0 & 0.0 & 1000 & 0.0 & 0.0 & 1000 & 0.0 & 0.0 & 1000 & 0.0 & 0.0 & 1000 & 0.0 & 0.0 & 1000 \\
\midrule

\multicolumn{16}{l}{\textbf{Hidden Object}} \\
Grim Tales: The Bride & 0.0 & 0.0 & 1000 & 0.0 & 83.3 & 1000 & 0.0 & 0.0 & 1000 & 0.0 & 0.0 & 1000 & 0.0 & 75.0 & 1000 \\
Grim Tales: The Legacy Collector’s Edition & 0.0 & 0.0 & 1000 & 0.0 & 75.0 & 1000 & 0.0 & 16.7 & 1000 & 0.0 & 0.0 & 1000 & 0.0 & 83.3 & 1000 \\
\midrule

\multicolumn{16}{l}{\textbf{Room Escape}} \\
Computer Office Escape & 0.0 & 0.0 & 1000 & 0.0 & 0.0 & 1000 & 0.0 & 0.0 & 1000 & 0.0 & 0.0 & 1000 & 0.0 & 0.0 & 1000 \\
Crimson Room & 0.0 & 14.3 & 1000 & 0.0 & 28.6 & 1000 & 0.0 & 0.0 & 1000 & 0.0 & 14.3 & 1000 & 0.0 & 0.0 & 1000 \\
Camping Room Escape & 0.0 & 22.2 & 1000 & 0.0 & 22.2 & 1000 & 0.0 & 11.1 & 1000 & 0.0 & 11.0 & 1000 & 0.0 & 22.2 & 1000 \\
Chemical Room Escape & 0.0 & 25.0 & 1000 & 0.0 & 25.0 & 1000 & 0.0 & 0.0 & 1000 & 0.0 & 12.5 & 1000 & 0.0 & 0.0 & 1000 \\
Space Museum Escape & 0.0 & 0.0 & 1000 & 0.0 & 0.0 & 1000 & 0.0 & 0.0 & 1000 & 0.0 & 0.0 & 1000 & 0.0 & 0.0 & 1000 \\
Vending Machine Room & 0.0 & 22.2 & 1000 & 0.0 & 11.1 & 1000 & 0.0 & 0.0 & 1000 & 0.0 & 11.1 & 1000 & 0.0 & 11.1 & 1000 \\
Wood Workshop Escape & 0.0 & 14.3 & 1000 & 0.0 & 28.6 & 1000 & 0.0 & 0.0 & 1000 & 0.0 & 0.0 & 1000 & 0.0 & 0.0 & 1000 \\
Geometric Room Escape & 0.0 & 0.0 & 1000 & 0.0 & 0.0 & 1000 & 0.0 & 0.0 & 1000 & 0.0 & 0.0 & 1000 & 0.0 & 0.0 & 1000 \\
Game Cafe Escape & 0.0 & 0.0 & 1000 & 0.0 & 0.0 & 1000 & 0.0 & 0.0 & 1000 & 0.0 & 0.0 & 1000 & 0.0 & 0.0 & 1000 \\
Machine Room Escape & 0.0 & 0.0 & 1000 & 0.0 & 0.0 & 1000 & 0.0 & 0.0 & 1000 & 0.0 & 0.0 & 1000 & 0.0 & 0.0 & 1000 \\
VideoStudio Escape & 0.0 & 0.0 & 1000 & 0.0 & 0.0 & 1000 & 0.0 & 0.0 & 1000 & 0.0 & 0.0 & 1000 & 0.0 & 0.0 & 1000 \\
Design House Escape & 0.0 & 0.0 & 1000 & 0.0 & 0.0 & 1000 & 0.0 & 0.0 & 1000 & 0.0 & 0.0 & 1000 & 0.0 & 0.0 & 1000 \\
Paint Room Escape & 0.0 & 0.0 & 1000 & 0.0 & 0.0 & 1000 & 0.0 & 0.0 & 1000 & 0.0 & 0.0 & 1000 & 0.0 & 0.0 & 1000 \\
Mirror Room Escape & 0.0 & 0.0 & 1000 & 0.0 & 0.0 & 1000 & 0.0 & 0.0 & 1000 & 0.0 & 0.0 & 1000 & 0.0 & 0.0 & 1000 \\
Elevator Room Escape & 0.0 & 0.0 & 1000 & 0.0 & 0.0 & 1000 & 0.0 & 0.0 & 1000 & 0.0 & 0.0 & 1000 & 0.0 & 0.0 & 1000 \\
\midrule

\multicolumn{16}{l}{\textbf{Visual Novel (Dating Sim)}} \\
Pico Sim Date & 0.0 & 0.9 & 1000 & 0.0 & 1.1 & 1000 & 0.0 & 0.0 & 1000 & 0.0 & 0.5 & 1000 & 0.0 & 0.0 & 1000 \\
Festival Days Sim Date & 0.0 & 10.6 & 1000 & 0.0 & 0.6 & 1000 & 0.0 & 0.0 & 1000 & 0.0 & 5.4 & 1000 & 0.0 & 8.5 & 1000 \\
Kingdom Days & 0.0 & 4.5 & 1000 & 0.0 & 4.5 & 1000 & 0.0 & 0.0 & 1000 & 0.0 & 1.5 & 1000 & 0.0 & 3.0 & 1000 \\
Idol Days Sim Date & 0.0 & 23.3 & 1000 & 0.0 & 11.6 & 1000 & 0.0 & 0.0 & 1000 & 0.0 & 17.4 & 1000 & 0.0 & 7.0 & 1000 \\
\midrule

\multicolumn{16}{l}{\textbf{Simulation}} \\
Sort the Court & 0.0 & 67.0 & 1000 & 0.0 & 54.4 & 1000 & 0.0 & 12.8 & 1000 & 0.0 & 64.6 & 1000 & 0.0 & 11.6 & 1000 \\
Community College Sim & 0.0 & 2.0 & 1000 & 0.0 & 1.9 & 1000 & 0.0 & 0.2 & 1000 & 0.0 & 2.4 & 1000 & 0.0 & 1.3 & 1000 \\
\midrule
\textbf{Average} & 0.0 & 6.6 & 1000 & 0.0 & 10.6 & 1000 & 0.0 & 1.2 & 1000 & 0.0 & 4.6 & 1000 & 0.0 & 6.9 & 1000 \\
\bottomrule
\end{tabular}
}
\caption{Detailed results for all 34 games across five GUI agents (\ie (1) Claude-3.7-Sonnet + UGround-V1-7B / \texttt{pyautogui} + Cradle, (2) Claude-3.7-Sonnet + Claude-3.7-Sonnet / \texttt{pyautogui} + Cradle, (3) Claude-3.7-Sonnet + Claude-3.7-Sonnet / \texttt{pyautogui} + Agent S2, (4) GPT-4o + UGround-V1-7B / \texttt{pyautogui} + Cradle, and (5) UI-TARS-1.5-7B).}
\label{tab:detailed_results_13}
\end{table*}

\subsection{Experimental Results}
\label{subsec:appendix_experimental_results}
We report the full experimental results on all \numGames games using OpenAI CUA, Claude-3.7-Sonnet Computer-Use, and ours with human performance in Table~\ref{tab:detailed_results_34}.
In addition, we report the results of other GUI agents in Table~\ref{tab:detailed_results_13}.

Overall, despite having the smallest parameter size, UI-TARS-1.5-7B achieves a higher milestone completion rate than proprietary LLM-based models like GPT-4o + UGround-V1-7B + Cradle and Claude-3.7-Sonnet + Claude-3.7-Sonnet + Agent S2, somehow consistent with its original claim of effectively solving browser-based games.
UGround-V1-7B, which incorporates a GUI grounding module, demonstrated weaker grounding ability compared to Claude-3.7-Sonnet.
Finally, although Agent S2 performs well in desktop GUI tasks~\citep{xie2024osworld}, it shows the lowest performance among the compared agents on \benchmarkName, highlighting the gap between everyday digital interactions and the complexities of adventure games.

\begin{table*}[h!]
\centering
\begin{adjustbox}{width=\linewidth}
\begin{tabular}{llcccc}
\toprule
\textbf{Game (Subgenre)} & \textbf{Metric} & 
\makecell{\textbf{GPT-4o} \\ \textbf{+ UGround} \\ \textbf{+ Cradle}} & 
\makecell{\textbf{Claude-3.7} \\ \textbf{+ Claude-3.7} \\ \textbf{+ Cradle}} & 
\makecell{\textbf{Claude-3.7} \\ \textbf{Computer-Use}} & 
\makecell{\textbf{Claude-3.7} \\ \textbf{Computer-Use} \\ \textbf{+ COAST}} \\
\midrule
\multirow{3}{*}{\makecell[l]{Sherlock Holmes 2 (Mystery/Detective)}} 
  & Success (\%)   & 0.0 $\pm$ 0.0   & 0.0 $\pm$ 0.0   & 0.0 $\pm$ 0.0   & 0.0 $\pm$ 0.0 \\
  & Milestone (\%) & 4.2 $\pm$ 7.2   & 16.7 $\pm$ 7.2  & 37.5 $\pm$ 0.0  & 54.2 $\pm$ 7.2 \\
  & \# Steps       & 1000.0 $\pm$ 0.0 & 1000.0 $\pm$ 0.0 & 1000.0 $\pm$ 0.0 & 1000.0 $\pm$ 0.0 \\
\midrule
\multirow{3}{*}{\makecell[l]{Grim Tales: The Bride (Hidden Object)}}
  & Success (\%)   & 0.0 $\pm$ 0.0   & 0.0 $\pm$ 0.0   & 0.0 $\pm$ 0.0   & 100.0 $\pm$ 0.0 \\
  & Milestone (\%) & 0.0 $\pm$ 0.0   & 75.0 $\pm$ 8.3  & 83.3 $\pm$ 8.3  & 100.0 $\pm$ 0.0 \\
  & \# Steps       & 1000.0 $\pm$ 0.0 & 1000.0 $\pm$ 0.0 & 1000.0 $\pm$ 0.0 & 281.3 $\pm$ 61.6 \\
\midrule
\multirow{3}{*}{\makecell[l]{Camping Room Escape (Room Escape)}}
  & Success (\%)   & 0.0 $\pm$ 0.0   & 0.0 $\pm$ 0.0   & 0.0 $\pm$ 0.0   & 0.0 $\pm$ 0.0 \\
  & Milestone (\%) & 3.7 $\pm$ 6.4   & 7.4 $\pm$ 12.8  & 25.9 $\pm$ 6.4  & 40.7 $\pm$ 6.4 \\
  & \# Steps       & 1000.0 $\pm$ 0.0 & 1000.0 $\pm$ 0.0 & 1000.0 $\pm$ 0.0 & 1000.0 $\pm$ 0.0 \\
\midrule
\multirow{3}{*}{\makecell[l]{Idol Days Sim Date (Visual Novel)}}
  & Success (\%)   & 0.0 $\pm$ 0.0   & 0.0 $\pm$ 0.0   & 0.0 $\pm$ 0.0   & 0.0 $\pm$ 0.0 \\
  & Milestone (\%) & 17.1 $\pm$ 4.8  & 10.1 $\pm$ 2.7  & 79.1 $\pm$ 35.8 & 45.3 $\pm$ 36.3 \\
  & \# Steps       & 1000.0 $\pm$ 0.0 & 1000.0 $\pm$ 0.0 & 920.0 $\pm$ 138.6 & 1000.0 $\pm$ 0.0 \\
\midrule
\multirow{3}{*}{\makecell[l]{Sort the Court (Simulation)}}
  & Success (\%)   & 0.0 $\pm$ 0.0   & 0.0 $\pm$ 0.0   & 33.3 $\pm$ 57.7 & 66.7 $\pm$ 57.7 \\
  & Milestone (\%) & 55.9 $\pm$ 8.0  & 53.9 $\pm$ 0.4  & 83.4 $\pm$ 7.1  & 101.2 $\pm$ 6.2 \\
  & \# Steps       & 1000.0 $\pm$ 0.0 & 1000.0 $\pm$ 0.0 & 990.7 $\pm$ 16.2 & 973.0 $\pm$ 23.5 \\
\midrule
\multirow{3}{*}{\textbf{Total}}
  & Success (\%)   & 0.0 $\pm$ 0.0   & 0.0 $\pm$ 0.0   & \underline{6.7 $\pm$ 11.5}  & \textbf{33.3 $\pm$ 11.5} \\
  & Milestone (\%) & 16.2 $\pm$ 2.0  & 32.6 $\pm$ 4.4  & \underline{62.0 $\pm$ 4.4}  & \textbf{68.3 $\pm$ 7.2} \\
  & \# Steps       & 1000.0 $\pm$ 0.0 & 1000.0 $\pm$ 0.0 & 982.1 $\pm$ 26.2 & 850.9 $\pm$ 8.6 \\
\bottomrule
\end{tabular}
\end{adjustbox}
\caption{Comparison of GUI agents in multi-run experiments on five sampled games. Values represent mean $\pm$ standard deviation.}
\label{tab:results_multi_run}
\end{table*}

\subsubsection{Multi-Run Experiments}
We additionally conducted multi-run experiments with three trials, and the results are shown in Table~\ref{tab:results_multi_run}. 
Claude-3.7 Computer-Use + COAST achieves a higher SR and MCR than baseline agents. 
These trends are consistent with the single-run results in Table~\ref{tab:performance_comparison_full}.

An exception occurs in \textit{Sort the Court}, where both agents complete the full story arc (SR > 0\%), suggesting stronger performance in NPC conversations (social reasoning) and resource management. 
However, this result is expected, as COAST already achieved a near-perfect MCR (95.5\%) in Table~\ref{tab:performance_comparison_full}.

MCR values can exceed 100\% in visual novel and simulation sub-genres because the continuous score is normalized relative to human performance. 
For instance, an MCR of 101.2\% in \textit{Sort the Court} indicates that the agent performs on par with, or even better than, a human player.

\begin{table*}[h!]
\centering
\begin{adjustbox}{width=\linewidth}
\begin{tabular}{llccc}
\toprule
\textbf{Game (Subgenre)} & \textbf{Metric} & 
\makecell{\textbf{Claude-3.7} \\ \textbf{Computer-Use}} & 
\makecell{\textbf{Claude-3.7} \\ \textbf{Computer-Use} \\ \textbf{(prompt-optimized)}} & 
\makecell{\textbf{Claude-3.7} \\ \textbf{Computer-Use} \\ \textbf{+ COAST}} \\
\midrule
\multirow{3}{*}{\makecell[l]{Sherlock Holmes 2 (Mystery/Detective)}}
  & Success (\%)   & 0.0   & 0.0   & 0.0   \\
  & Milestone (\%) & 37.5  & 37.5  & 62.5  \\
  & \# Steps       & 1000  & 1000  & 1000  \\
\midrule
\multirow{3}{*}{\makecell[l]{Grim Tales: The Bride (Hidden Object)}}
  & Success (\%)   & 0.0   & 0.0   & 100.0 \\
  & Milestone (\%) & 91.7  & 75.0  & 100.0 \\
  & \# Steps       & 1000  & 1000  & 225   \\
\midrule
\multirow{3}{*}{\makecell[l]{Camping Room Escape (Room Escape)}}
  & Success (\%)   & 0.0   & 0.0   & 0.0   \\
  & Milestone (\%) & 22.2  & 22.2  & 44.4  \\
  & \# Steps       & 1000  & 1000  & 1000  \\
\midrule
\multirow{3}{*}{\makecell[l]{Idol Days Sim Date (Visual Novel)}}
  & Success (\%)   & 0.0   & 0.0   & 0.0   \\
  & Milestone (\%) & 69.8  & 17.4  & 25.6  \\
  & \# Steps       & 760   & 1000  & 1000  \\
\midrule
\multirow{3}{*}{\makecell[l]{Sort the Court (Simulation)}}
  & Success (\%)   & 0.0   & 0.0   & 0.0   \\
  & Milestone (\%) & 78.1  & 80.8  & 95.5  \\
  & \# Steps       & 1000  & 1000  & 1000  \\
\midrule
\multirow{3}{*}{\textbf{Total}}
  & Success (\%)   & 0.0   & 0.0   & \textbf{20.0}  \\
  & Milestone (\%) & \underline{59.8}  & 46.6  & \textbf{65.6}  \\
  & \# Steps       & 952   & 1000  & 845   \\
\bottomrule
\end{tabular}
\end{adjustbox}
\caption{Comparison of Claude-3.7 Sonnet Computer-Use variants (\ie baseline, prompt-optimized, and \methodName) across five sampled games.}
\label{tab:claude_cua_variants}
\end{table*}

\subsubsection{Fair Comparison between \methodName and Baselines}
To demonstrate that the success of \methodName stems from its novel Seek-Map-Solve architecture rather than improved prompt engineering, we conducted an additional control experiment in which the baseline agent (\ie Claude-3.7 Computer-Use) was instructed using \methodName-style prompts in Appendix~\ref{subsubsec:appendix_detailed_prompts}.
Specifically, we concatenated prompts from the Clue Seeker, Clue Mapper, and Problem Solver modules, made minor adjustments for coherence, and then provided the final combined prompt to the Claude-3.7 Computer-Use.

Table~\ref{tab:claude_cua_variants} shows that prompt-optimized one is slightly worse than the naive Claude-3.7 Computer-Use.
The prompt-optimized version fails to solve additional milestones in the mystery/detective and room escape subgenres involving long-term observation-behavior gaps, where \methodName demonstrates clear gains.
These findings support our claim that \methodName’s performance improvements stem not from better prompting, but from its structured Seek-Map-Solve framework.

\begin{table*}[h!]
\scriptsize
\centering
\begin{tabularx}{\textwidth}{@{} p{0.05\textwidth} @{\hspace{0.5em}} p{0.2\textwidth} @{\hspace{1em}} >{\raggedright\arraybackslash}X @{\hspace{1em}} >{\raggedright\arraybackslash}X @{}}
\toprule
\textbf{ID} & \textbf{Game} & \textbf{Question} & \textbf{Expected Answer} \\
\midrule

Q1 & Sherlock Holmes: The Tea Shop Murder Mystery & 
Is there anything of interest inside the insurance agency? If so, what should the player do there to progress in the game? &
There are two employees and a bird in the room. The player should talk to the employees to obtain new suspects and can replace the newspapers in the birdcage. \\

\addlinespace[0.5em]
Q2 & Sherlock Holmes: The Tea Shop Murder Mystery & 
When the player arrives at Main Street, please describe exactly what they should do, including interactions, items to obtain, and how those items contribute to progressing the story. &
Talk to the boy who is selling newspapers to get one, and unlock new suspects. \\

\addlinespace[0.5em]
Q3 & Grim Tales: The Legacy Collector’s Edition & 
What should the player do when they reach the barrels on the board? Please describe any required interactions. &
The player should check the item list at the bottom of the screen and search for the hidden items within the game scene. \\

\addlinespace[0.5em]
Q4 & Grim Tales: The Legacy Collector’s Edition & 
What should the player find items at barrels on the board? List up please. &
Bee, Dwarf, Pepper, Clower, Star, Shell, Crab, Tomato, Train, Darts, Mushroom, and Machete \\

\addlinespace[0.5em]
Q5 & Grim Tales: The Legacy Collector’s Edition & 
Where is the bee at the barrels on the board? &
The bee is sitting on the rock directly above the toad, in the center of the screen. \\

\addlinespace[0.5em]
Q6 & Camping Room Escape & 
What color is the tent? Can I enter it? If I can, what item can I find inside? &
The tent is yellow-green. Yes, you can enter and find a sheet of paper. \\

\addlinespace[0.5em]
Q7 & Camping Room Escape & 
There is a case secured with a padlock. What is the combination to unlock it, and what happens after the case is opened? &
The code is 6428. It opens to get cardboard tubes. \\

\addlinespace[0.5em]
Q8 & Pico Sim Date & 
There’s a place where you can buy drinks. What happens when you do? Does it affect any of your stats or relationships? &
Buying drinks increases charm. \\

\addlinespace[0.5em]
Q9 & Pico Sim Date & 
When the player goes to work at the game, what actions do they perform, and what are the outcomes or benefits? &
You can work like a mailroom tech and earn money from it. If your intelligence score is above a certain threshold, you have the chance to get promoted. (Note: You can attempt promotion only once per day.) \\

\addlinespace[0.5em]
Q10 & Community College Sim & 
Which role should the player choose to start the game with Job Level 2 already unlocked? &
Choose the "Adult" role. \\

\bottomrule
\end{tabularx}
\caption{Questions for contamination check with expected answers.}
\label{tab:Questions for contamination check}
\end{table*}

\section{Further Analysis}
\subsection{Analysis on Contaminations}
\label{subsec:appendix_contamination}

\begin{table*}[h!]
\centering
\scriptsize
\resizebox{\linewidth}{!}{%
\begin{tabular}{ll|l|l|l|l|l|l|l|l|l|l}
\hline
\multicolumn{2}{l|}{LLM}                       & Q1 & Q2     & Q3 & Q4 & Q5 & Q6     & Q7 & Q8 & Q9     & Q10 \\ \hline
\multirow{3}{*}{GPT-4o}            & Correct    & \xmark  & $\sim$ & \xmark  & \xmark  & \xmark  & $\sim$ & \xmark  & \xmark  & $\sim$ & \xmark   \\
                                   & Conclusion & \xmark  & \xmark      & \xmark  & \xmark  & \xmark  & \cmark      & \xmark  & \xmark  & \xmark      & \xmark   \\
                                   & Behavior   & \cmark  & \cmark      & \cmark  & -  & \cmark  & \xmark      & \xmark  & -  & \cmark      & -   \\ \hline
\multirow{3}{*}{Claude-3.7-Sonnet} & Correct    & \xmark  & \xmark      & \xmark  & \xmark  & \xmark  & \xmark      & \xmark  & \xmark  & \xmark      & \xmark   \\
                                   & Conclusion & \xmark  & \xmark      & \xmark  & \xmark  & \xmark  & \xmark      & \xmark  & \xmark  & \xmark      & \xmark   \\
                                   & Behavior   & \xmark  & \cmark      & \cmark  & -  & \cmark  & \xmark      & \xmark  & -  & \cmark      & -   \\ \hline
\end{tabular}%
}
\caption{Evaluation results for contamination check questions. Responses that are only partially correct or contain hallucinated content are marked with a `$\sim$' symbol. If the agent does not encounter a situation where the relevant knowledge is needed, it is marked with a `-'.}
\label{tab:Results of contamination check}
\end{table*}

For the contamination check, we composed 1 to 3 questions per game(5 games) and assessed the extent of contamination in each LLM model through human evaluation. The specific questions and their expected answers are provided in Table~\ref{tab:Questions for contamination check}. The evaluation criteria are as follows:
\begin{enumerate}
    \item \textbf{Answer Accuracy}: How well does the response align with information from the walkthrough?
    \item \textbf{Conclusion Accuracy}: Is the judgement about the outcome in each response correct?
    \item \textbf{Behavior Match}: Does the agent's actual behavior during evaluation align with the correct standard?
\end{enumerate}

The results of this evaluation for each questions are presented in Table~\ref{tab:Results of contamination check}. These results show the contamination observed in several questions for GPT-4o. For example, in Question 2, the model correctly identified that there is a boy on the street and that a newspaper can be obtained. However, instead of stating that the newspaper should be acquired from the boy, the model hallucinated alternative actions-such as purchasing it from a store or suggesting other implausible interactions-indicating that its knowledge was partially contaminated or imprecise.

\subsection{Additional Hint Injection}
\label{subsec:appendix_additional_hint_injection}

To enable hint injection, we constructed hints from complete walkthroughs by decomposing them into subtasks aligned with each game’s milestone structure. Rather than embedding these hints in the initial prompt, we adopted a conditional injection strategy: hints were introduced only when the model failed to make progress toward a milestone over a predefined number of steps.

Specifically, in \textit{Sherlock Holmes: The Tea Shop Murder Mystery}, a hint was injected if no milestone progress was made after 100 steps. Without hints, the model achieved only a single milestone across 1,000 turns. In contrast, with hint injection, it successfully completed all milestones within 758 turns. In \textit{Computer Office Escape}, hints were provided every 50 turns. Despite offering guidance on relevant patterns and subsequent tasks in Phase 1, the model failed to achieve even the first milestone.

The hints used for each game and the final prompt with injected hints are summarized below:

\begin{tcolorbox}[title=Hints for ``Sherlock Holmes: The Tea Shop Murder Mystery'',
colback=gray!10, colframe=green!50!black, breakable]
\small
\textbf{Phase 1}
\begin{itemize}
  \item Go to Main Street and talk to the newspaper vendor. He will give you a new suspect.
  \item Get the following items at the tea shop: a gold key and an insurance document.
\end{itemize}

\textbf{Phase 2}
\begin{itemize}
  \item Go to McTavish Insurance agency. Talk with people in there. You can get some suspects.
  \item Give the newspaper to bird.
\end{itemize}

\textbf{Phase 3}
\begin{itemize}
  \item Go to Wellington Inn and give sugar at hole. After that, talk with people in there. You can get some suspects.
\end{itemize}

\textbf{Phase 4}
\begin{itemize}
  \item Go to McTavish Insurance agency and give duster to Sena.
  \item Go to the tea shop, put record on the gramophone.
\end{itemize}

\textbf{Phase 5}
\begin{itemize}
  \item Use crab to cut the chain at the tea shop.
  \item Enter the door which is opened by cutting the chain.
\end{itemize}

\textbf{Phase 6}
\begin{itemize}
  \item Enter the door which is opened by cutting the chain at the tea shop and click the portrait.
  \item Use the clues and the suspect note to select the culprit.
\end{itemize}

\end{tcolorbox}

\begin{tcolorbox}[title=Hints for ``Computer Office Escape'',
colback=gray!10, colframe=green!50!black, breakable]
\small
\textbf{Phase 1}
\begin{itemize}
  \item To unlock the computer on the bookshelf, observe the pattern of the books and click the matching pattern on the computer.
  \item Find the orange blind. Look at the pattern on it and use that to unlock the gray box beneath it.
\end{itemize}

\textbf{Phase 2}
\begin{itemize}
  \item Go to the desk with the quarter-circle shape. Use the circular patterns on the desks to unlock the pattern on the computer.
  \item Connect the two items you’ve obtained to the computer to rotate it.
\end{itemize}

\textbf{Phase 3}
\begin{itemize}
  \item Use the red USB you found to open a drawer.
\end{itemize}

\textbf{Phase 4}
\begin{itemize}
  \item Place the cup you obtained on the coffee machine. When it’s filled with coffee, drink it.
  \item Use the pattern on the bottom of the coffee cup to unlock the box directly below the coffee machine.
\end{itemize}

\textbf{Phase 5}
\begin{itemize}
  \item Use the brown key to unlock the closed door.
\end{itemize}

\textbf{Phase 6}
\begin{itemize}
  \item On top of the dollar-sign cabinet, there’s a gray box. Look at the pattern on the sticky note next to it, and press the buttons on the tablet on the floor to receive a code.
  \item Input the code into the gray box on top of the dollar-sign cabinet.
\end{itemize}

\textbf{Phase 7}
\begin{itemize}
  \item Place the object with red squares on top of the computer keyboard with the green USB.
  \item From the revealed information, enter the letters in alphabetical order into the computer.
\end{itemize}

\textbf{Phase 8}
\begin{itemize}
  \item Obtain the tablet with the blue USB inserted.
\end{itemize}

\textbf{Phase 9}
\begin{itemize}
  \item Find the sign with the rotating arrow.
  \item Use this to solve the similar pattern on the computer.
\end{itemize}

\textbf{Phase 10}
\begin{itemize}
  \item Find the drawer that can be opened with the green USB.
  \item Using the information from the tablet and the computer in the room with the orange blinds, solve the puzzle on the newly obtained black tablet.
\end{itemize}

\textbf{Phase 11}
\begin{itemize}
  \item Use the black tablet to activate a flashlight. Shine it on the sunflower.
  \item Memorize the directions shown and find the flower-patterned box. Enter the directions in order.
\end{itemize}

\textbf{Phase 12}
\begin{itemize}
  \item Find the box that can be opened with the silver key. It’s located beyond the brown door.
  \item Inside the opened box, look at the color pattern of the electric wires and use it to solve the problem on the tablet labeled A B C D outside.
\end{itemize}

\textbf{Phase 13}
\begin{itemize}
  \item Check the email on the black tablet.
  \item To enter the password, match each symbol with the corresponding table shape and input accordingly.
\end{itemize}

\textbf{Phase 14}
\begin{itemize}
  \item The QR code from the tablet is only half. Find the missing half and place the tablet there.
  \item Use the tablet with the blue USB to scan the completed QR code, as it has the scanning function.
\end{itemize}

\textbf{Phase 15}
\begin{itemize}
  \item Find the drawer that can be opened with the blue USB.
  \item Use a screwdriver to open the hub on the computer labeled “39.”
\end{itemize}

\textbf{Phase 16}
\begin{itemize}
  \item Use the 3 USBs and the hub you obtained to open the final door.
\end{itemize}

\end{tcolorbox}

\begin{tcolorbox}[title=Hint Injection Prompt for ``Sherlock Holmes: The Tea Shop Murder Mystery'',
colback=gray!10, colframe=green!50!black, breakable]
\small
\textbf{[Instruction]} \\
You’re a game agent solving an adventure game. Adventure games can involve variables and often require lateral thinking and creative problem-solving. Rather than focusing solely on the ultimate goal, try to solve problems using common sense and imaginative thinking.

\vspace{0.5em}
\textbf{[Important Notice]} \\
To ensure a smooth experience and prevent unexpected issues, please read and follow the instructions below carefully:
\begin{itemize}
    \item Do not repeatedly interact with objects that may not be clickable. Even if nothing happens, your action might have already been registered, or the object may simply not be interactive.
    \item Avoid repeating the same action excessively. A lack of visible response or an unexpected result does not mean the action failed.
    \item Do not click the setup or question buttons in the top-left corner of the screen.
    \item Do not leave the game screen.
    \item Do not access Settings or Help:
    \begin{itemize}
        \item Do not click the wrench icon (Settings).
        \item Do not click the question mark icon (Help or Hints).
    \end{itemize}
\end{itemize}

\vspace{0.5em}
\textbf{[Prompt]} \\
You are now the lead agent in ``Sherlock Holmes: The Tea Shop Murder Mystery'', a point-and-click detective game. Everything in this game is controlled with simple mouse clicks. Search the tea shop carefully, interact with objects, uncover hidden clues, and solve the murder mystery.

\vspace{0.5em}
\textbf{[Basic Rules]}
\begin{enumerate}
    \item Observe carefully. Don’t rush or search randomly. Every detail could matter. What seems ordinary might hide secrets.
    \item Think logically and laterally. Don’t brute-force or guess wildly. Follow the trail of clues using reason, insight, and creativity.
    \item Investigate step by step. Whether solving a mystery or unlocking a puzzle, progress comes from careful observation and deduction.
    \item Trust your instincts. Sometimes the answer is hidden in plain sight. Don’t overlook the obvious.
\end{enumerate}

\vspace{0.5em}
\textbf{[Information]}
\begin{itemize}
    \item Mr. T was murdered by someone. But we don't know who the criminal is.
    \item Mr. T was the owner of the local tea shop.
    \item You must find the true criminal of this case.
    \item You can find items at crime scenes and collect clues by traveling to different locations using the map.
    \item Clues you find in the environment will be added to the inventory at the bottom of the screen. To use an item, click on it and place it in the appropriate location.
\end{itemize}

\vspace{0.5em}
\textbf{[Completion Condition]} \\
The game is considered complete only when you have gathered enough evidence, logically identified the culprit, and correctly selected the true suspect during the final decision. A clear message or scene will confirm that the case has been solved.

\vspace{0.5em}
\textbf{[Completion Signal]} \\
Once you’ve solved the case, type the following to signal completion: \texttt{[Done]}

\vspace{0.5em}
\textbf{[Hint]} 
\begin{itemize}
    \item Go to Main Street and talk to the newspaper vendor. He will give you a new suspect.
    \item Get the following items at the tea shop: a gold key and an insurance document.
\end{itemize}

\end{tcolorbox}

\subsection{Large Reasoning Models}
\label{subsec:large_reasoning_models}
To evaluate whether a reasoning-specialized model improves action planning, we replaced GPT-4o + UGround-V1-7B + Cradle with o4-mini~\citep{openai2025o4mini} + UGround-V1-7B + Cradle, which is designed for stronger high-level reasoning.

In \textit{Sherlock Holmes: The Tea Shop Murder Mystery}, o4-mini achieved 0 milestones, and in \textit{Camping Room Escape}, it achieved 2: showing no improvements over GPT-4o’s results.
While it generated more complex and structured reasoning, it did not improve action planning.

\end{document}